\documentclass[a4paper,11pt,preprint,sort&compress,twocolumn,table]{elsarticle}

\journal{~}

\usepackage{subcaption}
\usepackage{amsmath}
\usepackage{amssymb}
\usepackage{tikz}
\usepackage{graphicx}
\usetikzlibrary{bayesnet, arrows}

\usepackage[margin=18mm]{geometry}
\usepackage[parfill]{parskip}

\usepackage{microtype}
\usepackage[hidelinks]{hyperref}
\usepackage[noabbrev]{cleveref}

\usepackage{amsmath,amssymb,amsfonts}
\usepackage{algorithmic}
\usepackage{float}
\usepackage{textcomp}
\usepackage{blindtext}

\usepackage{booktabs,tabulary,tabularx,array, makecell}
\newcolumntype{L}[1]{>{\centering\let\newline\\\arraybackslash\hspace{0pt}}m{#1}}

\usepackage{microtype}

\begin{document}

\twocolumn[{\begin{frontmatter}


\title{\textbf{Hierarchical Bayesian Modelling for Knowledge Transfer Across Engineering Fleets via Multitask Learning}}

    \author[1]{L.A.\ Bull \corref{cor1}}
    \ead{lbull@turing.ac.uk}
    \author[1]{D.\ Di Francesco}
    \author[2]{M.\ Dhada}
    \author[3]{O.\ Steinert}
    \author[4]{T.\ Lindgren}
    \author[2]{A.K.\ Parlikad}
    \author[1,5]{A.B.\ Duncan}
    \author[1,6]{M.~Girolami}

    \cortext[cor1]{Corresponding author}
    
    \address[1]{The Alan Turing Institute, The British Library, London, NW1 2DB, UK}
    \address[2]{Institute for Manufacturing, Department of Engineering, University of Cambridge, CB3 0FS, UK}
    \address[3]{Strategic Product Planning and Advanced Analytics, Scania CV, Scania AB (publ), SE-151 87 Södertälje, Sweden}
    \address[4]{Department of Computer and Systems Sciences, Stockholm University, P.O. Box 7003, SE-164 07 Kista, Sweden}
    \address[5]{Department of Mathematics, Imperial College London, London, SW7 2AZ, UK}
    \address[6]{Department of Engineering, University of Cambridge, CB3 0FA, UK}

\begin{abstract}

A population-level analysis is proposed to address data sparsity when building predictive models for engineering infrastructure. %
Utilising an interpretable hierarchical Bayesian approach and operational fleet data, domain expertise is naturally encoded (and appropriately shared) between different sub-groups, representing (i) use-type, (ii) component, or (iii) operating condition. %
Specifically, domain expertise is exploited to constrain the model via assumptions (and prior distributions) allowing the methodology to automatically share information between similar assets, improving the survival analysis of a truck fleet 
and power prediction in a wind farm. 
In each asset management example, a set of correlated functions is learnt over the fleet, in a combined inference, to learn a population model. %
Parameter estimation is improved when sub-fleets 
share correlated information at different levels of the hierarchy. 
In turn, groups with incomplete data automatically borrow statistical strength from those that are data-rich. %
The statistical correlations enable knowledge transfer via Bayesian transfer learning, and the correlations can be inspected to inform which assets share information for which \textit{effect} (i.e.\ parameter). 
Both case studies demonstrate the wide applicability to practical infrastructure monitoring, since the approach is naturally adapted between interpretable fleet models of different \textit{in situ} examples. %

\end{abstract}


\end{frontmatter}
}]

\section{Introduction}

Data sparsity can cause significant issues in practical applications of reliability, performance, and safety assessment. %
Particularly structural monitoring~\cite{worden2007application}, prognostics \cite{o2012practical}, or performance and health management \cite{kim2017prognostics}. %
In these domains, comprehensive (or high variance \cite{paleyes2020challenges}) data are rarely available \textit{a priori}; instead, measurements arrive incrementally, throughout the life-cycle of the monitored system~\cite{bull2019probabilistic}. %
For example, the data recorded from the system in unusual environments, or following damage, might take years to collect. %
Labelling to annotate the measurements can also be limited or expensive, requiring input from a domain expert. %
Such incomplete data motivate \textit{sharing} information between similar assets; specifically, whether systems with comprehensive data (or established models) can support those with incomplete information. %

The concept of knowledge transfer, from one machine to another, has led to the development of population-based~\cite{bull2021foundations, gosliga2021foundations, gardner2021foundations} or fleet monitoring \cite{zaccaria2018fleet}. %
Initial investigations (mostly) consider the quantification of \textit{similarity} between systems~\cite{gosliga2021foundations} and tools for the \textit{transfer} of data and/or models from \textit{source} to \textit{target} domains~\cite{michau2019domain,bull2021transfer,gardner2021overcoming}. %
An alternative approach is considered here, whereby a combined inference is made given the measurements from a collected group of systems~\cite{dhada2020anomaly}. %
Specifically, a set of correlated, hierarchical models is learnt, given the information recorded from the collected population. %
Two case studies are presented: survival analysis of an operational truck fleet and wind-power predictions for an operational wind farm. %
Population-level models are learnt using hierarchical Bayesian modelling~\cite{wand2009semiparametric,gelman2013bayesian} providing robust predictions and variance reduction compared to independent models 
and two benchmarks. %
The \textit{multi-task learning} approach \cite{murphy2012machine,wand2009semiparametric} automatically shares information between correlated domains (i.e.\ sub-groups) such that assets with sparse information borrow statistical strength from those that are data-rich (via correlated variables). %

\subsection{Why learn fleet models?}\label{s:why-EngTL}
Throughout this work, the term \textit{fleet} refers to a population of assets that constitute engineering infrastructure. %
For example, civil structures (bridges and roads) or vehicles (trains in a rail network). %
The problem setting from each case study is introduced here to motivate multitask learning from \textit{in-situ} fleet data. %

\subsubsection*{Truck fleets}

The first example concerns the survival analysis of components (alternators and turbochargers) in a fleet of heavy-duty trucks maintained by Scania~CV. %
The components are maintained in a run-to-failure strategy as failure models are unavailable and it is infeasible for drivers to sense incipient failure. %
Nonetheless, the associated downtime can incur high costs: relating to late goods delivery, re-loading, and towing vehicles to the workshop. %

For such components, survival analysis \cite{o2012practical} is critical to estimate the time to failure, and therefore fundamental when designing a maintenance plan. %
The analysis considers failure occurrences in the population over some specified time period. %
The period must be sufficiently long, such that reliability functions can be evaluated based on observed failures or drop-outs \cite{birolini2013reliability}. %
Specifically, this work focusses on the hazard function $\lambda(t)$ which defines the instantaneous rate of failure -- it is the probability $P(\cdot)$ of a component failing at time $t$, given that it has survived until time $t$~\cite{birolini2013reliability}, %
\begin{equation}
\lambda(t) = \frac{P(t \leq T < t + dt | T \geq t)}{dt}
\end{equation}
here $T$ denotes the \textit{time of failure}. %
Empirically, this is calculated as the fraction of trucks that failed to the number of trucks that survived, in a given time interval. %

Importantly, each sample from the reliability function requires at least one failure in the historical fleet data. %
For this reason, if failures are rare in certain sub-fleets, the data that represent the corresponding function will be sparse -- \Cref{fig:truck-data} later visualises this. %
If sub-fleets with more failures can inform predictions in groups where failures are rare, this greatly extends the value of the measured data (and the failure events themselves). %

\subsubsection*{Wind farms}

The second case study considers power prediction for a group of operational wind turbines. %
Here, the regression tasks 
are \textit{power curves}, which map from wind speed to power output for a specific turbine~\cite{papatheou2017performance}. %
The associated function can be used as an indicator of performance and is useful in monitoring procedures~\cite{ROGERS20201124}. %
Data-based methods approximate this relationship from operational measurements, typically recorded using Supervisor Control and Sensory Data Acquisition (SCADA) systems~\cite{YANG2013365}. %
Various techniques have been proposed to model data that correspond to \emph{normal} operation~\cite{thapar2011critical,carrillo2013review,lydia2014comprehensive}. %
In practice, however, only a subset of measurements represent this relationship. %
In particular, power \textit{curtailments} will appear as additional functional components; %
these usually correspond to the output power being controlled (or otherwise limited) by the operator.
Reasons for this action include: adhering to the requirements of the electrical grid \cite{waite2016modeling,hur2014curtailment}, the mitigation of loading/wake effects~\cite{bontekoning2017analysis}, and restrictions enforced by planning regulations~--~such data are presented in \Cref{f:PC-data}. %

Critically, different turbines experience different conditions (i.e.\ power curves) at varying intervals. %
If the power of a particular turbine is regularly limited by the operator (as a result of its location in the farm) measurements collected from this operation become far more valuable when they can be shared between turbines. %
In this case, fleet modelling can be adopted to share (or \textit{pool}) information. 

\subsection{Novelty}

In view of these applications, the proposed fleet modelling approach favours explainability (with some caveats) since each model is informative. %
\begin{itemize}
    \item Rather than black-box, a fleet model is built while encoding multilevel \textit{a priori} knowledge of fleet behaviour and model constraints, given domain expertise. %
    \item The proposed model automatically determines the level of knowledge transfer between asset groups, learning the inter-task correlations from data and combining this with \textit{a priori} engineering knowledge. %
    \item In turn, the approach provides formal uncertainty quantification of the fleet effects (parameters) at various asset group granularities (system-specific, operating condition, or population-wide). %
    \item Each subgroup predictor shares information and the associated \textit{fleet model} provides new insights, which are greater than the sum of its parts (single-task learning). %
\end{itemize}

Such fleet models are desirable, since they enable downstream analyses, to determine which groups of assets share information for which (interpretable) parameter; additionally, the model naturally integrates with experimental design or decision processes; formalising the expected optimal action, or the value data collection activities -- these concepts are demonstrated in the second case study, \Cref{s:decisions}. %

The approach is particularly suited to sparse, incremental data, that are found in many (practical) monitoring applications -- for example, in the first (survival analysis) case study, one domain owns a single training observation. %

\subsection{Layout}
The paper layout is as follows. %
\Cref{s:related-work} summarises existing work relating to population monitoring of engineering systems. %
\Cref{s:cont} states the contributions of this work. %
\Cref{s:HBMEM} introduces a general methodology for knowledge transfer via hierarchical Bayesian modelling. %
\Cref{s:trucks,s:wind-farms} present the truck fleet and wind farm case studies. \Cref{s:conc} offers concluding remarks. %

\section{Related Work}\label{s:related-work}

A summary of fleet-monitoring literature is provided. %
The term \textit{knowledge transfer} is used generally to refer to methods that learn from multiple related datasets. 
Specific definitions of \textit{transfer learning} are contentious: this work follows \citet{murphy2012machine} which views multitask learning (MTL) as the combined inference of a set of related tasks, while domain adaptation (DA) is a method of transforming data, such that the same task can be learnt for multiple domains. %
Both approaches are considered \textit{transfer learning} -- especially when domains share interpretable, parametrised models. %

\subsection{Fine-tuning and domain adaptation}
When monitoring engineering populations, the majority of literature focusses on \textit{transfer learning}. %
Transfer learning seeks to improve predictions in a \textit{target domain} given the information in a (more complete) \textit{source domain}. %
Many examples consider crack detection via image classification using Convolutional Neural Networks (CNNs). %
For example, \citet{dorafshan2018comparison,gao2018deep, jang2019deep} detect cracks over a number of domains by \textit{fine-tuning} the parameters of a CNN trained on a source domain to aid generalisation in the target. %

Domain Adaptation is viewed as another variant of transfer learning in engineering applications (DA)~\cite{zhang2017new,li2019multi,wang2019domain}. %
These techniques define some mapping from domain data into a shared space (possibly one of the original domains) where a \textit{single} model is used to make predictions. %
For example, \citet{michau2019domain} apply a neural network mapping for DA in the condition monitoring of a fleet of power plants. %
DA has also been investigated by (kernelised) linear projection, discussed in a structural health monitoring context by \citet{gardner2020application, gardner2020machine} considering methods for knowledge transfer between simulated source and target structures, as well as a simulated source and experimental target structure~\cite{gardner2022application}. %
Damage detectors have also been transferred between systems via DA in a group of tailplane structures using ground-test vibration data~\cite{bull2021transfer}. %
To accommodate for class imbalance and data sparsity, often associated with monitoring data, \citet{pool22} introduce statistic alignment methods for adaptation procedures. %

\subsection{Multi-task learning}
An alternative view of population-level models considers multi-task learning (MTL). %
While the multi-task approach also assumes the predictors (i.e.\ \textit{tasks}) are correlated over the fleet, the parameters across domains are learnt at the same time with equal importance. %
A combined inference allows \textit{domain-specific} models to share information between related tasks, improving the accuracy in domains where data are limited~\cite{sun2021vector}. %

Examples of multi-task learning are less prevalent when modelling engineering infrastructure. %
\citet{wan2019bayesian} successfully use a Gaussian process (GP) to learn correlations between tasks in a multi-output regression. %
The GP is built using a carefully specified kernel \cite{bonilla2007multi} to capture the task and inter-task relationships. %
The experiments capture correlations between temperature/acceleration sensing systems on a single structure (the Canton Tower), rather than multiple assets in a fleet. %
Similarly, \citet{li2021missing} apply correlated GPs to address the missing data problem over multiple sensors of a hydroelectric dam. 
The results demonstrate successful knowledge transfer between measurement channels. %
Considering aerospace engines, \citet{seshadri2020bayesian} apply GPs for knowledge transfer between multiple axial measurement planes when interpolating temperature fields within an aircraft engine. 
Sharing information between planes significantly improves the spatial representation of the response. %

Hierarchical Bayesian modelling offers another multi-task framework. %
A model is built with a `hierarchy' of parameters, whereby domain-specific tasks are correlated via \textit{shared} latent variables (explained in \Cref{s:HBMEM}). %
The approach was introduced to structural monitoring by \citet{huang2019multitask} and \citet{huang2015hierarchical} who utilise hierarchical models to learn multiple, correlated regression models for modal analysis. %
A shared sparseness profile is inferred over all tasks and related measurement channels, improving damage detection and data recovery by considering the correlation between damage scenarios or adjacent sensors on the same structure. %
Some recent, related applications include \citet{di2021decision}, who use hierarchical models to build corrosion models given evidence from multiple locations, and \citet{papadimas2021hierarchical}, where the results from a series of materials experiments (i.e.\ coupon samples) are combined to inform the estimation of material properties. %
Also, \citet{dhada2020anomaly} implement hierarchical Gaussian mixture models to cluster simulated data that represent novelty detection for asset management; the model parameters are interpretable in terms of the data distribution, rather than the application domain. %

\subsection{Wider monitoring methods}

It is worth considering more general developments in the literature, and how they relate to fleet monitoring. %
Multi-task neural networks, in particular, show promise when the size (or features) of monitoring data permit their application; e.g.\ \citet{zhang2020multi} design a deep architecture for guided wave datasets. %
Similarly, \citet{tsialiamanis2022application} successfully investigate neural networks for knowledge transfer by mapping measurements from multiple structures onto a common manifold, to learn a shared representation. %


A primary motivation of this work, however, is to consider structures/domains with very sparse (or absent) data -- e.g.\ those recently in operation, or new environmental conditions. %
In turn, model comparisons here are limited to parametric (or \textit{shallow}  \cite{sukhija2020shallow}) methods of knowledge transfer, centred around interpretable models -- each benchmark is outlined in \Cref{s:EngApps}. %

\subsection{Bayesian vs `deep' knowledge transfer}

The distinction between \textit{hierarchical} (Bayesian) and \textit{deep} (neural network) approaches to transfer learning is important. %
The differences emphasise why, in many applications, the proposed (hierarchical) method is required for infrastructure monitoring. %

\begin{itemize}
    \item Both address relative data sparsity (between domains) however, the level of sparsity is method dependent: generally, deep methods are suited to complex features and big data; hierarchical methods are suited to standard measurements 
    and interpretable models.
    \item Both improve predictions over multiple asset groups; however, the proposed hierarchical approach provides uncertainty quantification of the nested subgroups, enabling downstream (statistical) analyses -- e.g.\ experimental design or decision processes (demonstrated in \Cref{s:decisions}).
    \item Encoding domain (engineering) expertise is natural for multilevel Bayesian models -- for example, the knowledge that all turbines in a wind farm have the same maximum power, but the rate at which they limit to a maximum will depend on turbine location.
    \item Conversely, for neural networks, encoding domain expertise is difficult since they are nonparametric; in turn, the inferences (and model constraints) at different levels of fleet granularity are less intuitive.
\end{itemize}


\section{Contribution}\label{s:cont}
The main contributions of this work are twofold: (i) multi-task learning with hierarchical Bayesian modelling allows information to be shared between distinct (but related) systems using operational fleet data (wind turbines and trucks) rather than multiple sensors on a single structure; (ii) various \textit{mixed effects} are considered in the hierarchy, such that certain characteristics (parameters) can be learnt at the individual, group, or population level. %
In turn, prior engineering knowledge can be encoded at different levels in the hierarchy and parameters can be shared for various (nested) subgroups. 
The hierarchical models are easily formulated around interpretable parameters and the resultant structure allows insightful analyses of the predicted variables, indicating which groups of systems share information for which effect. 

When multi-task learning for engineered infrastructure, it is crucial to establish an appropriate level of knowledge transfer (data pooling) between systems or domains. %
If information is inappropriately shared, this can lead to \textit{negative transfer}, whereby population models prove worse than conventional (single task) learning. %
Importantly, the proposed model automatically determines an appropriate level of knowledge transfer, by learning the inter-task correlations from the data and combining this with engineering knowledge -- encoded as prior distributions within the hierarchical structure. %

The resultant approach permits formal uncertainty quantification at various levels of the predictive model, and, in turn, various granularities of fleet behaviour (e.g.\ system-specific, condition-specific, or population-wide). %
Multiple levels of uncertainty quantification enable natural integration with decision processes, or experimental design procedures, considering the whole fleet. %
In turn, the model can be used to inform fleet interactions within a wider asset management programme. %
To highlight this novelty, the hierarchical model is integrated with a demonstrative decision process in the second (wind farm) case study. %

Similarly, while the proposed hierarchical model makes inferences from observations at the sub-fleet level only (i.e.\ task-specific outputs) predictions can be made at various levels -- including larger groups and the aggregated population. %
Inference of the joint population model (from task-specific observations) presents the knowledge transfer mechanism. %
The resultant structure produces both \textit{shared} and \textit{task-specific} models -- this is not true for any of the benchmarks, which learn one of the two options (i.e.\ single-task learning, complete pooling, domain adaptation -- \Cref{s:EngApps}). %

\section{Hierarchical Bayesian Modelling for Multi-Task Learning with Mixed Effects}\label{s:HBMEM}
Consider fleet data, recorded from a population of engineering systems, which are separated into $K$ groups or \textit{sub-fleets}. %
The population data can then be denoted,
\begin{align}
\left\{\mathbf{x}_k, \mathbf{y}_k\right\}_{k=1}^K = \left\{\left\{x_{ik}, y_{ik}\right\}_{i=1}^{N_k}\right\}_{k=1}^K
\end{align}

where $\mathbf{y}_{k}$ is target response vector for inputs $\mathbf{x}_{k}$ and $\{x_{ik}, y_{ik}\}$ are the $i^{th}$ pair of observations in group $k$. %
There are $N_k$ observations in each group and thus $\sum_{k=1}^{K} N_k$ observations in total. %
The aim is to learn a set of $K$ predictors, one for each group, related to classification or regression tasks. %
Without loss of generality, this work focusses on the regression setting, where the tasks satisfy, 
$$
\left\{ y_{ik} = f_k(x_{ik}) + \epsilon_{ik} \right\}_{k=1}^K
$$
i.e.\ the output is determined by evaluating one of $K$ latent functions with additive noise $\epsilon_{ik}$. %
Note, for classification, logistic regression would involve modifying the above likelihood for categorisation (a Bernoulli distribution) and passing $f_k(x_{ik})$ through the logit function to ensure predictions are between zero and one (binary classification)~\cite{murphy2012machine}. %

The mapping $f_k$ is assumed to be correlated between sub-fleets. %
In consequence, the models should be improved by learning the parameters in a joint inference over the whole population. %
In machine learning, this is referred to as \textit{multi-task learning}; in statistics, such data are usually modelled with hierarchical models \cite{kreft1998introducing, gelman2006data}. %

\subsection{Hierarchical Bayesian modelling}
In practice, while certain sub-fleets might have rich, historical data, others (particularly those recently in operation) will have limited training data. %
In this setting, learning separate, independent models for each group will lead to unreliable predictions. %
On the other hand, a single regression of all the data (complete pooling) will result in poor generalisation. %
Instead, hierarchical models can be used to learn separate models for each group while encouraging task parameters to be correlated~\cite{murphy2012machine} -- %
the established theory is summarised here.

Consider $K$ linear regression models, 
\begin{align}
\bigg\{ \mathbf{y}_k = \boldsymbol{\Phi}_k \boldsymbol{\alpha}_k + \boldsymbol{\epsilon}_k \bigg\}_{k=1}^K \label{e:i-model-eqns}
\end{align}
where $\boldsymbol{\Phi}_k = [\mathbf{1}, \mathbf{x}_k]$ is the $N_k  \times 2$ \textit{design} matrix; $\boldsymbol{\alpha}_k$ is the $2 \times 1$ vector of \textit{weights}; and the noise vector is $N_k \times 1$ and normally distributed\footnote{In this first introductory example, the additive noise variance $\sigma_k^2$ is observed -- in the next example, it is unobserved.} %
 $\boldsymbol{\epsilon}_k \sim \textrm{N}\left(0, \sigma_k^2\mathbf{I}\right)$. %
$\mathbf{1}$ is a vector of ones, $\mathbf{I}$ is the identity matrix, and $\textrm{N}(m, s)$ is the normal distribution with mean $m$ and (co)variance $s$.
The likelihood of the target response vector is then,
\begin{align}
\mathbf{y}_{k}| \mathbf{x}_{k} &\sim \textrm{N}\left(\mathbf{\Phi}_k \boldsymbol{\alpha}_k,\; \sigma^2_k \mathbf{I}\right) \\
\therefore \quad y_{ik}| x_{ik} &\sim \textrm{N}\left(\alpha^{(k)}_1 + \alpha^{(k)}_2 x_{ik},\; \sigma^2_k\right) \nonumber
\end{align}

In a Bayesian manner, one can set a shared hierarchy of prior distributions over the weights %
(slope and intercept) for the groups %
$k \in \{1,\ldots,K\}$,
\begin{align}
\{\boldsymbol{\alpha}_k\}_{k=1}^{K} &\overset{\textrm{\tiny i.i.d}}{\sim} \textrm{N}\left(\boldsymbol{\mu}_{\alpha}, \textrm{diag}\left\{\boldsymbol{\sigma}^2_{\alpha}\right\}\right) \label{e:prior_alpha}\\
\boldsymbol{\mu}_{\alpha} &\sim \textrm{N}\left(\textbf{m}_\alpha, \textrm{diag}\left\{\textbf{s}_\alpha\right\}\right) \label{e:prior_mu_alpha}\\
\boldsymbol{\sigma}_{\alpha} &\overset{\textrm{\tiny i.i.d}}{\sim} \textrm{IG}(a, b) \label{e:prior_sig_alpha}
\end{align} 
In words, (\ref{e:prior_alpha}) assumes that the weights $\{\boldsymbol{\alpha}_k\}_{k=1}^K$ are normally distributed $\textrm{N}(\cdot)$ with mean $\boldsymbol{\mu}_{\alpha}$ and covariance\footnote{The operator $\textrm{diag}\{\mathbf{a}\}$ forms a square diagonal matrix with the elements from $\mathbf{a}$ on the main diagonal and zeros elsewhere.} $\textrm{diag}\{\boldsymbol{\sigma}^2_{\alpha}\}$. %
Similarly,~(\ref{e:prior_mu_alpha}) states that the prior expectation of the weights $\boldsymbol{\alpha}_k$ is normally distributed with mean $\textbf{m}_\alpha$ and covariance $\textrm{diag}\{\textbf{s}_\alpha\}$; %
(\ref{e:prior_sig_alpha}) states that the prior deviation of the slope and intercept is inverse-Gamma distributed $\textrm{IG}(\cdot)$ with shape and scale parameters $a$ and $b$ respectively. %

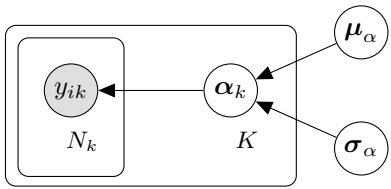
\begin{figure}[ht]
  \centering
  \begin{tikzpicture}
    \node[obs] (y) {$y_{ik}$};%
    \node[latent,right=of y,xshift=1em] (alpha) {$\boldsymbol{\alpha}_k$}; %
    \node[latent,right=of alpha,yshift=2em] (am) {$\boldsymbol{\mu}_{\alpha}$};
    \node[latent,right=of alpha,yshift=-2em] (as) {$\boldsymbol{\sigma}_{\alpha}$};
    \plate [inner sep=3ex] {plate2} {(y)(alpha)} {$K$}; %
    \plate [inner sep=2ex] {plate1} {(y)} {$N_k$}; %
    \edge {alpha} {y}
    \edge {am,as} {alpha}
    \end{tikzpicture}
    \caption{DGM of hierarchical linear regression.}\label{f:hb-lr}
\end{figure}

Selecting appropriate prior distributions, and their associated hyperparameters $\{\textbf{m}_\alpha,\textbf{s}_\alpha, a, b \}$, is essential to the success of hierarchical models. %
In this work, prior elicitation is justified by encoding engineering knowledge in each case study as weakly informative priors~\cite{gelman2013bayesian}. %
The Directed Graphical Model (DGM) in \Cref{f:hb-lr} visualises the general hierarchical regression. %
The nodes show observed/latent variables as shaded/non-shaded respectively; arrows show conditional dependencies, and plates show multiple instances of sub-scripted nodes.

The $K$ weight vectors $\boldsymbol{\alpha}_k$ are correlated via the common latent variables $\{\boldsymbol{\mu}_{\alpha}, \boldsymbol{\sigma}^2_{\alpha}\}$; i.e.\ parent nodes in \Cref{f:hb-lr}. %
Note that \cref{e:prior_alpha,e:prior_mu_alpha,e:prior_sig_alpha} encode \textit{prior} belief of the independence between latent variables. %
In this work, this does not restrict the covariance structure of the posterior distribution for $\{\boldsymbol{\alpha}_k\}_{k=1}^K$ since it is approximated using Markov Chain Monte Carlo (MCMC, summarised in \Cref{s:inference}). %

Via correlations in the posterior distribution, sparse domains borrow statistical strength from those that are data-rich. %
Crucially, to \textit{share} information between tasks, the parent nodes $\{\boldsymbol{\mu}_{\alpha}, \boldsymbol{\sigma}^2_{\alpha}\}$ must be inferred from the population data. %
In this way, the sub-fleet parameters $\boldsymbol{\alpha}_k$ are (indirectly) influenced by the wider population. %
Consider that, if $\{\boldsymbol{\mu}_{\alpha}, \boldsymbol{\sigma}^2_{\alpha}\}$ were fixed constants, rather than variables inferred from data, each model would be conditionally independent, preventing the \textit{flow} of information between domains~\cite{murphy2012machine}. %

\subsection{Mixed-effects modelling}
The hierarchical structure allows \textit{effects} (i.e.\ interpretable latent variables) to be learnt at different levels, as well as `prior' information. %
Specifically, the parameters of the model itself (\ref{e:i-model-eqns}) can be learnt at the system, sub-fleet, or population level. %
The inference of parameters at various levels of hierarchy, while encoding engineering/domain knowledge at each level, constitutes significant novelty here. %

Returning to the regression example (\ref{e:i-model-eqns}), consider that the variance $\sigma^2_k$ of the noise $\boldsymbol{\epsilon}_k$ is in fact unknown. %
While one could learn $K$ domain-specific noise variance terms $\sigma^2_k$, it is typically assumed that the noise is equivalent across tasks. %
Sharing the parameter and inferring it from the population can significantly reduce the uncertainty in its prediction. %
Of course, this assumption should be justified given an understanding of the problem at hand; for example, the same sensing system collects all the population data. %
In terms of notation, (\ref{e:i-model-eqns}) remains the same, however, the domain-specific noise vector $\boldsymbol{\epsilon}_k$ is now distributed %
$\boldsymbol{\epsilon}_k \sim \textrm{N}\left(0, \sigma^2\mathbf{I}\right)$.
The removal of subscript-$k$ from the noise variance implies that the size of $\sigma^2$ remains the same while the number of the sub-fleets $K$ increases (unlike $\boldsymbol{\alpha}_k$). %
Intuitively, $\sigma^2$ is now a \textit{tied} parameter~\cite{murphy2012machine}. %

Similarly, it makes sense to also infer \textit{effects} at the population level, to further reduce model uncertainty%
\footnote{For example, the intercept would be a shared parameter, with zero-mean, in a related linear regression of Hooke's law for several materials tests.}. %
Throughout this work, it is assumed that \textit{shared} effects also enter the model linearly, for the target response vector~$\mathbf{y}_{k}$ and inputs~$\mathbf{x}_{k}$,
\begin{align}
\bigg\{ \mathbf{y}_k = \underbrace{\boldsymbol{\Phi}_k \boldsymbol{\alpha}_k}_{\textrm{random}} + \underbrace{\boldsymbol{\Psi}_k \boldsymbol{\beta}}_{\textrm{fixed}} + \, \boldsymbol{\epsilon}_k \bigg\}_{k=1}^K \label{e:mem-model-eqns}
\end{align}
Where $\boldsymbol{\Psi}_k$ is some design matrix of inputs, and $\boldsymbol{\beta}$ is the corresponding vector of weights. %
Again, there is no subscript-$k$ for $\boldsymbol{\beta}$ (like $\sigma^2$) as it is tied between sub-fleets. %
Following~\citet{kreft1998introducing}, the $\boldsymbol{\beta}$ coefficients as considered \textit{fixed effects}, as they are learnt at the population level and shared, while $\boldsymbol{\alpha}_k$ are \textit{random effects}, as they vary between \textit{individuals}. %
Intuitively, a model with both fixed and random effects can be considered a \textit{mixed} (effects) model \cite{west2006linear,gelman2013bayesian}. %
\Cref{{f:hb-me-lr}} shows the modified DGM of the hierarchical regression. %
The key differences are nodes outside of the $K$ plate -- these are the tied parameters, learnt at the population level. %

\begin{figure}[ht]
  \centering
  \begin{tikzpicture}
    \node[obs] (y) {$y_{ik}$};%
    \node[latent,right=of y] (alpha) {$\boldsymbol{\alpha}_k$}; %
    \node[latent,left=of y] (beta) {$\boldsymbol{\beta}$}; %
    \node[latent,above=of y, yshift=-1ex] (sigma) {$\sigma$}; %
    \node[latent,right=of alpha,yshift=2em] (am) {$\boldsymbol{\mu}_{\alpha}$};
    \node[latent,right=of alpha,yshift=-2em] (as) {$\boldsymbol{\sigma}_{\alpha}$};
    \plate [inner sep=3ex] {plate2} {(y)(alpha)} {$K$}; %
    \plate [inner sep=2ex] {plate1} {(y)} {$N_k$}; %
    \edge {sigma,alpha,beta} {y}
    \edge {am,as} {alpha}
    \end{tikzpicture}
    \caption{
    DGM of hierarchical linear regression with mixed effects. %
    }\label{f:hb-me-lr}
\end{figure}
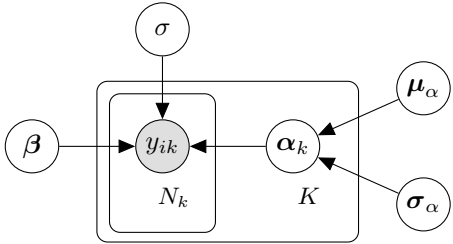

As \citet{gelman2013bayesian} point out, the terms \textit{random} and \textit{fixed} originate from a frequentist perspective and are somewhat confusing in a Bayesian context where all parameters are random, or (equivalently) fixed with unknown values. %
The terminology is used, however, as it is intuitive considering engineering applications and consistent with established literature in modelling panel or longitudinal data \cite{gelman2006data}. %
One should also consider that interpreting mixed-effects models remains challenging, even when models are parametrised. %
If the effects are not (linearly) independent, the fixed and random coefficients can influence each other, making it difficult to reliably recover their relationships. %
In turn, the modelling assumptions must be carefully considered when emphasising interpretability. %

\subsection{Inference}\label{s:inference}
In view of graphical models, the observed variables are referred to as \textit{evidence} nodes. %
For example, the hierarchical regression in \Cref{f:hb-lr} would have the following set of evidence nodes, 
\begin{align}
    \mathcal{E} = \{[\mathbf{y}_k]\}
\end{align}
where $[\mathbf{y}_k]$ is shorthand to denote complete set $\{\mathbf{y}_1, \mathbf{y}_2, \ldots, \mathbf{y}_K\}$.
On the other hand, the latent variables are \textit{hidden} nodes,
\begin{align}
    \mathcal{H} = \{[\boldsymbol{\alpha}_k], \boldsymbol{\mu}_\alpha, \boldsymbol{\sigma}_\alpha\}
\end{align}
Bayesian inference relies on finding the posterior distribution of $\mathcal{H}$ given $\mathcal{E}$, i.e.\ the distribution of the unknown parameters given the data,
\begin{align}
    p(\mathcal{H}|\mathcal{E}) &= \frac{p(\mathcal{H},\mathcal{E})}{p(\mathcal{E})} \nonumber \\
    &= \frac{p([\mathbf{y}_k, \boldsymbol{\alpha}_k], \boldsymbol{\mu}_\alpha, \boldsymbol{\sigma}_\alpha)}{
    p(\mathbf{y}_k)}\nonumber\\
    = &\frac{p([\mathbf{y}_k]|[\boldsymbol{\alpha}_k])p([\boldsymbol{\alpha}_k] | \boldsymbol{\mu}_\alpha, \boldsymbol{\sigma}_\alpha)p( \boldsymbol{\mu}_\alpha) p(\boldsymbol{\sigma}_\alpha)}{
    \int \int \int p([\mathbf{y}_k, \boldsymbol{\alpha}_k], \boldsymbol{\mu}_\alpha, \boldsymbol{\sigma}_\alpha) \;
    d\boldsymbol{\alpha}_k d\boldsymbol{\mu}_\alpha d\boldsymbol{\sigma}_\alpha} \label{e:e_inference}
\end{align}

DGM representations are useful since inference can be aided by graph-theoretic results. %
The systematic application of graph-theoretic algorithms has led to a number of probabilistic programming languages~--~here, %
models are implemented in \texttt{Stan} \cite{carpenter2017stan}. %
The parameters are inferred using MCMC, via the no U-turn implementation of Hamiltonian Monte Carlo~\cite{hoffman2014no}. %
Throughout, the burn-in period is 1000 iterations and 2000 iterations are used for inference. %
Code based on the first case study is publicly available on \href{https://github.com/labull/EngineeringPatternRecognition}{GitHub}\footnote{Rather than the operational data presented here, the code uses simulated data (in view of data sensitivity).}.

\subsection{Engineering applications}\label{s:EngApps}

In each case study, hierarchical models are formulated for knowledge transfer between asset models. %
The first concerns survival analysis of truck fleets (hazard curves) and the second concerns power prediction for turbines  (power curves). %
Engineering expertise is encoded in a number of ways: to (i)~inform prior elicitation, (ii) determine which effects are random or fixed, and (ii) formulate interpretable parameters. %
In turn, population modelling offers insights as to which sub-fleets share information for which (interpretable) effect. %

Importantly, by considering the collected population, the training data can, in effect, be extended. %
In turn, parameter estimation is improved while increasing the reliability of predictions. %
There are, of course, important considerations when building such models -- prior elicitation, mixed-effects formulation, negative transfer -- these concepts are discussed throughout. %

Throughout, the predictive performance of the multitask methodology (MTL) is compared to three fleet monitoring benchmarks: %
\begin{itemize}
    \item (STL) Single Task Learning: the predictive model learnt from each domain independently.
    \item (CP) Complete pooling: the predictive model learnt from the collected fleet data, assuming all data are generated by a single task.
    \item (CRL) Correlation alignment for domain adaptation: sequentially treating each task $\hat{k}$ as the target domain, and embedding the remaining (source) domains onto the joint distribution $p(\mathbf{y}_{\hat{k}}, \mathbf{x}_{\hat{k}})$ using CORAL~\cite{sun2017correlation}. All measurements are treated as one task, and a single model is learnt, to predict the \textit{target} test data.
\end{itemize}

For sensible comparisons, the predictive model is consistent across all benchmarks -- what differs is the effective \textit{presentation} of data during inference. %
Note that parameter interpretation becomes problematic in domain adaptation (CRL) since the (source) joint distributions $\{p(\mathbf{y}_k, \mathbf{x}_k)\}_{k=1}^K$ have been transformed onto the target \cite{pool22} . %
Once transformed, making predictions for new source observations is nontrivial. %
These caveats highlight a benefit of the proposed methodology; %
however, comparisons to CRL are included to emphasise that adaptation alone is insufficient to treat all fleet monitoring problems, especially with parametrised models and sparse data. %

By nature of the practical applications (and data sensitivity) in each case study, validation to a ground truth for \textit{parameters} is not feasible; for this reason, models are compared to the available (response) ground truth and quantified by the predictive log-likelihood (e.g.\ (\ref{eq:post_pred1})). %

\section{Truck-Fleet Survival Analysis}\label{s:trucks}

\begin{figure}[t]
    \centering\includegraphics[width=.9\linewidth]{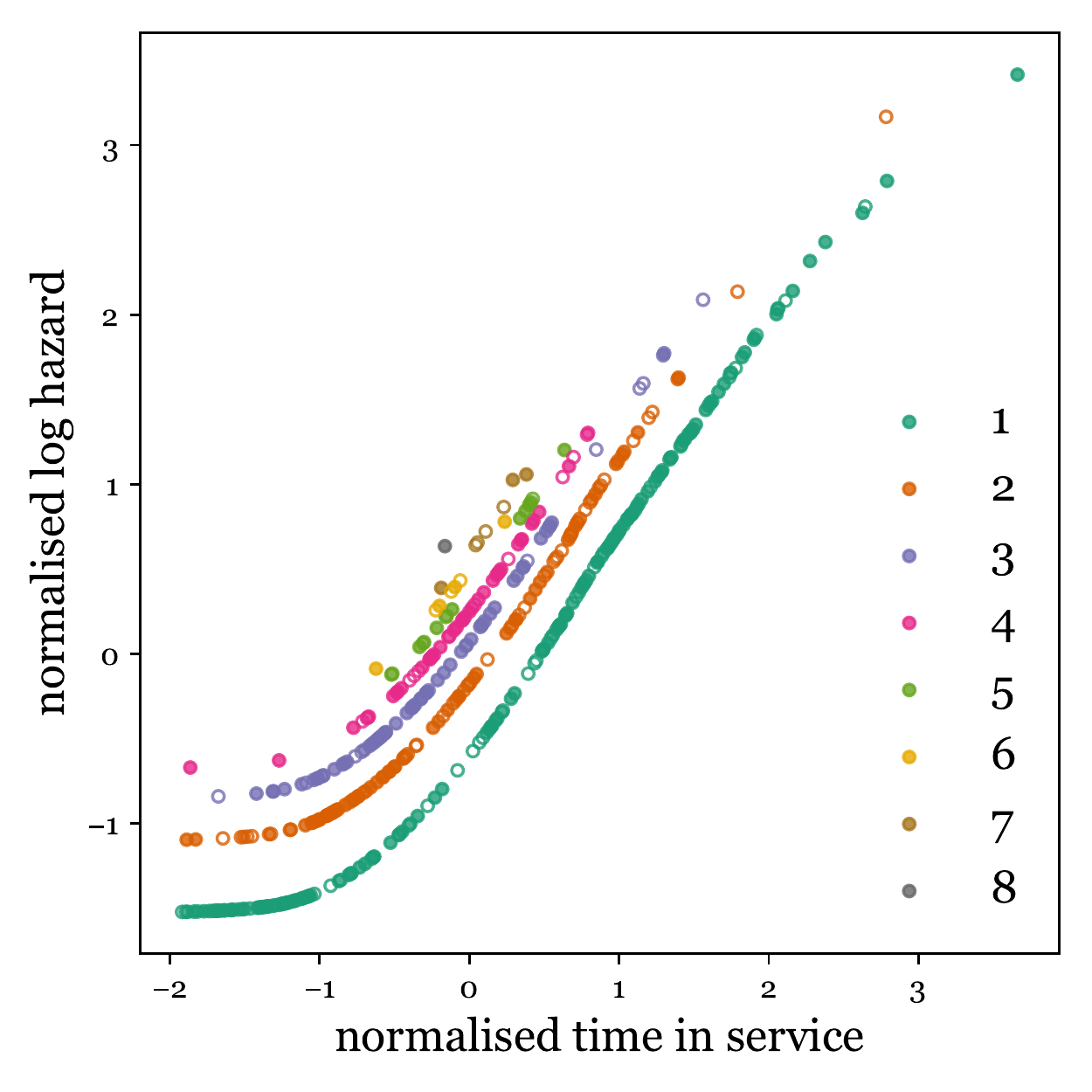}
    \caption{Log hazard function data for alternators in the truck fleet. Training and testing markers are $\bullet$ and $\circ$ respectively. Colours correspond to sub-fleet labels, associated with the task index $k \in \{1,2,\ldots, 8\}$. (\textit{\ref{a:post-pub} post-publication note}) }
    \label{fig:truck-data}
\end{figure}

The hazard data for truck fleet alternators are shown in \Cref{fig:truck-data}. %
Herein, this work considers the log-hazard, since it is easier to visualise. 
There are 437 observations in total, split into a 75\% training set and 25\% test set. %
The data are z-score normalised in view of data sensitivity and certain (specific) details are omitted. %
The observations represent the complete monitoring dataset, since no observations we lost via normalisation, truncation, or censorship. %
It is clarified that normalisation affects the \textit{direct} interpretation of the parameters. %
In practice, however, one can recover interpretable values by transforming back into the original space. %
Here, for the purpose of discussion, the relative parameter values and their relationships remain interpretable. %
To generate the hazard data, the total time in service for all assets was divided into intervals of one day; for each day, the ratio of the number of components that failed to the number that survived (so far) is calculated. %
The choice of interval length is dependent on the application -- here one day is sufficient compared to the maintenance horizon. %

The sub-fleets were manually labelled in collaboration with the engineers at Scania. %
Colours correspond to different sub-populations, where the total number of groups (and, therefore, hazard functions) is ${K=8}$. %
(\textit{\ref{a:post-pub} post-publication note}.) %
Note that certain domains are more sparse than others, with the most extreme case being $k=8$, owning a single observation. %
The population model will look to utilise data-rich domains with more information ($k \in \{1,2,3\}$) to support the sparse domains ($k \in \{5,6,7,8\}$). %
The number of task-wise observations is as follows,
\begin{center}
\begin{tabular}{| c | c | c | c | c | c | c | c || c |}
\hline
 $N_1$ & $N_2$ & $N_3$ & $N_4$ & $N_5$ & $N_6$ & $N_7$ & $N_8$ & $\sum_{k=1}^K N_k$\\
 \hline
 180 & 108 & 70 & 49 & 15 & 7 & 7 & 1 & 437 \\
 \hline
\end{tabular}
\end{center}

\subsection{Task regression formulation}\label{s:truck-model}

When analysing survival data, it is convenient to assume the survival time~$T$ is parametrically distributed since the parameters are interpretable and formulate a specific hazard function. %
A straightforward example is presented when $T$ is exponentially distributed, leading to a constant hazard~\cite{rodriguez2010parametric}. %

Rather than constant, \Cref{fig:truck-data} shows the log-hazard is near-linear for a large proportion of the input domain, with a notable nonlinear effect at low $t$ values (early hours in service). %
Therefore, it is assumed the best (parametric) approximation of the marginal $p(T=t)$ is the Gompertz distribution (G) for each sub-fleet~\cite{rodriguez2010parametric}, %
\begin{align}
p(T=t) &= \textrm{G}(t\; ; \;\gamma, \phi) \nonumber \\
& = (\gamma e^{\phi t}) \exp \left\{- \frac{\gamma}{\phi} (e^{\phi t} - 1)\right\} \label{e:GM}
\end{align}
This is convenient, since (\ref{e:GM}) is formulated such that log-hazard is linear in time~$t$,
\begin{align}
\log \lambda_{\textrm{G}}(t) &=  \log\gamma + \phi t \nonumber \\
& =  \alpha_1 + \alpha_2 t \label{e:gm-hazard}
\end{align}
Since only hazard data were available, tasks are fit directly to~(\ref{e:gm-hazard}) rather than the distribution over the time at failure (\ref{e:GM}). %
The correct likelihood, however, should consider the distribution (\ref{e:GM}) as the tasks directly -- this avoids assumptions of a Gaussian likelihood for the log-hazard. 
Instead, the (log) hazard uncertainty would be naturally represented by the variance of $\gamma$ and $\phi$. %
Unfortunately, this was not possible here in view of data availability. %
For a better interpretation of the parameters in practice, and agreement with Kolmogorov's axioms, the likelihood of the population model should represent the time-at-failure $T$ directly. %

Considering the data in \Cref{fig:truck-data}, a weighted sum of $H$ B-spline bases functions~$b_h(t)$ is included to model the (non-parametric) discrepancy between the linear Gompertz hazard and the empirical data,  %
\begin{align}
\log \lambda(t) &=  \alpha_1 + \alpha_2 t + \sum^H_{h=1} \beta_h b_h(t) \label{e:basis-funcs}\\
  & = \log \lambda_{\textrm{G}}(t) + \sum^H_{h=1} \beta_h b_h(t) \nonumber
\end{align}
Cubic B-splines (\ref{a:b-splines}) are selected as they are smooth with compact support, resulting in a sparse design matrix for the $\beta_h$ terms\footnote{An appropriate number of splines $H$ will be determined through cross-validation. $H$ is treated deterministically to simplify implementation and improve stability since the uncertainty of $H$ is less informative compared to more interpretable parameters}. %
This property is suitable since the nonlinear response acts in specific (compact) regions of the input. %
In effect, (\ref{e:basis-funcs}) defines a semi-parametric (or a partially-linear) regression \cite{wand2009semiparametric} with kernel smoothing to approximate the hazard functions for each sub-fleet. %

\subsection{Mixed-effects formulation}\label{s:truck-effects}
From \Cref{fig:truck-data}, one observes the underlying linear trend $\{\alpha_1 + \alpha_2 t\}$ is varying between sub-fleets while the nonlinear effect $\sum^H_{h=1} \beta_h b_h(t)$ appears consistent over the population. %
In other words, while the data are poorly described by a (linear) Gompertz hazard function, the (nonparametric) discrepancy remains consistent. %

Therefore, the associated spline weights $\boldsymbol{\beta} = \{\beta_h\}_{h=1}^H$ are assumed to be fixed effects and learnt at the population level. %
On the other hand, task-specific linear weights are inferred, which are correlated via common latent variables (random effects) $\boldsymbol{\alpha}_k = \{\alpha^{(k)}_1, \alpha^{(k)}_2\}$. %

The mixed effect model can now be expressed in the general notation from (\ref{e:mem-model-eqns}),
\begin{align}
\bigg\{ \mathbf{y}_k = \underbrace{\boldsymbol{\Phi}_k \boldsymbol{\alpha}_k}_{\textrm{random}} + \underbrace{\boldsymbol{\Psi}_k \boldsymbol{\beta}}_{\textrm{fixed}} + \, \boldsymbol{\epsilon}_k \bigg\}_{k=1}^K \nonumber 
\end{align}
Specifically, for each sub-fleet $k$: $\mathbf{y}_k$ is the output of the log-hazard (\ref{e:basis-funcs}) with additive noise $\boldsymbol{\epsilon}_k$; $\mathbf{x}_k$ are the inputs corresponding to time $t$; $\boldsymbol{\alpha}_k$ is the varying linear weight vector with design matrix ${\boldsymbol{\Phi}_k = [\mathbf{1}, \mathbf{x}_k]}$; and $\boldsymbol{\beta}$ is the tied/fixed weight vector, with a design matrix of splines, %
\begin{align}
    \boldsymbol{\Psi}_k = 
\begin{bmatrix} 
  b_1(\mathbf{x}_k),b_2(\mathbf{x}_k), \ldots, b_H(\mathbf{x}_k)
\end{bmatrix}
\end{align}
The resultant graphical model corresponds to \Cref{f:hb-me-lr} %
and the likelihood of the response is,
\begin{align}
y_{ik}| x_{ik}, \boldsymbol{\theta}_k& \sim \nonumber\\ 
&\textrm{N}\left(\alpha^{(k)}_1 + \alpha^{(k)}_2 x_{ik} + \sum^H_{h=1} \beta_h b_h(x_{ik}),\; \sigma^2\right) \label{eq:lik-1}
\end{align}

where $\boldsymbol{\theta}_k = \{\boldsymbol{\alpha}_k, \boldsymbol{\beta}, \boldsymbol{\mu}_\alpha, \boldsymbol{\sigma}_\alpha, \sigma \}$ is the set of parameters indexed to task $k$. %

\subsection{Weakly informative priors}

Primarily considering $\boldsymbol{\alpha}_k$, it is possible to encode prior knowledge of the expected functions, since the linear component corresponds to a Gompertz survival model (\ref{e:gm-hazard}). %
It is acknowledged that, in this case, the specific hyperparameter values are less meaningful as the data are normalised; however, their interpretation remains relevant. %

Specifically, $\boldsymbol{\alpha}_k$ is distributed according to \cref{e:prior_alpha,e:prior_mu_alpha,e:prior_sig_alpha}, with hyperparameters,
\begin{gather}
    \mathbf{m}_\alpha = [0, 1.5]^\top, \qquad \mathbf{s}_\alpha = [2, 0.5]^\top \label{e:prior-expect-alpha}\\
    a = 1, \qquad b = 1
\end{gather}
The first element of $\mathbf{m}_\alpha$ corresponds to the intercept and postulates the baseline log-hazard\footnote{Or the exponentiated initial rate-of-failure.}. (This is $0$ since the data are centred). %
The second element of $\mathbf{m}_\alpha$ is the expected slope of the log-hazard. %
(Set to $1.5$ as one expects hazard to increase exponentially under the Gompertz model with a gradient $>1$ when normalised). %
The $\mathbf{s}_\alpha$ values indicate a weakly informative prior under the ranges imposed by z-score normalisation. %
Similarly, the $a, b$ values encourage correlation between sub-fleet models, such that the prior mode of the standard deviation of the generating distribution of $\boldsymbol{\alpha}_k$ is $b/(a+1) = 1/2$ (this intentionally overestimates the deviation $\boldsymbol{\sigma}_{\alpha}$ between sub-fleets, such that the population model weakly constrains $\boldsymbol{\alpha}_k$). %

The shared prior over the variance of the additive noise $\boldsymbol{\epsilon}_k$ is set to,
\begin{align}
\sigma \sim \textrm{IG}(3, 0.8) \label{e:prior_simga}
\end{align}
Whose mode is at $0.2$, indicating that the standard deviation of the noise is expected to be significantly less (around five times) than that of the output, i.e.\ a high signal-to-noise ratio. %

Following a standard approach \cite{gelman2013bayesian} the basis function model can be centred around the linear component ($\log \lambda_{\textrm{G}}(t)$) via specification of the $\boldsymbol{\beta}$ prior. %
Specifically, one can postulate a \textit{shrinkage} prior with a high density at zero, to (effectively) exclude basis functions by encouraging their expected posterior weights to be near-zero -- while also having heavy tails to avoid over-shrinkage. %
A standard hierarchical prior is used \cite{tipping2001sparse} which exhibits these desired properties, %
\begin{align}
    \beta_h \sim \textrm{N}(0, \sigma_h^2), \qquad \sigma_h^2 \sim \textrm{IG}\left(v, v\right) \label{e:prior-beta}
\end{align}
where $v$ is some small nonzero value -- in this case $v = 10^{-3}$. %

\begin{figure}[t]
    \centering\includegraphics[width=.9\linewidth]{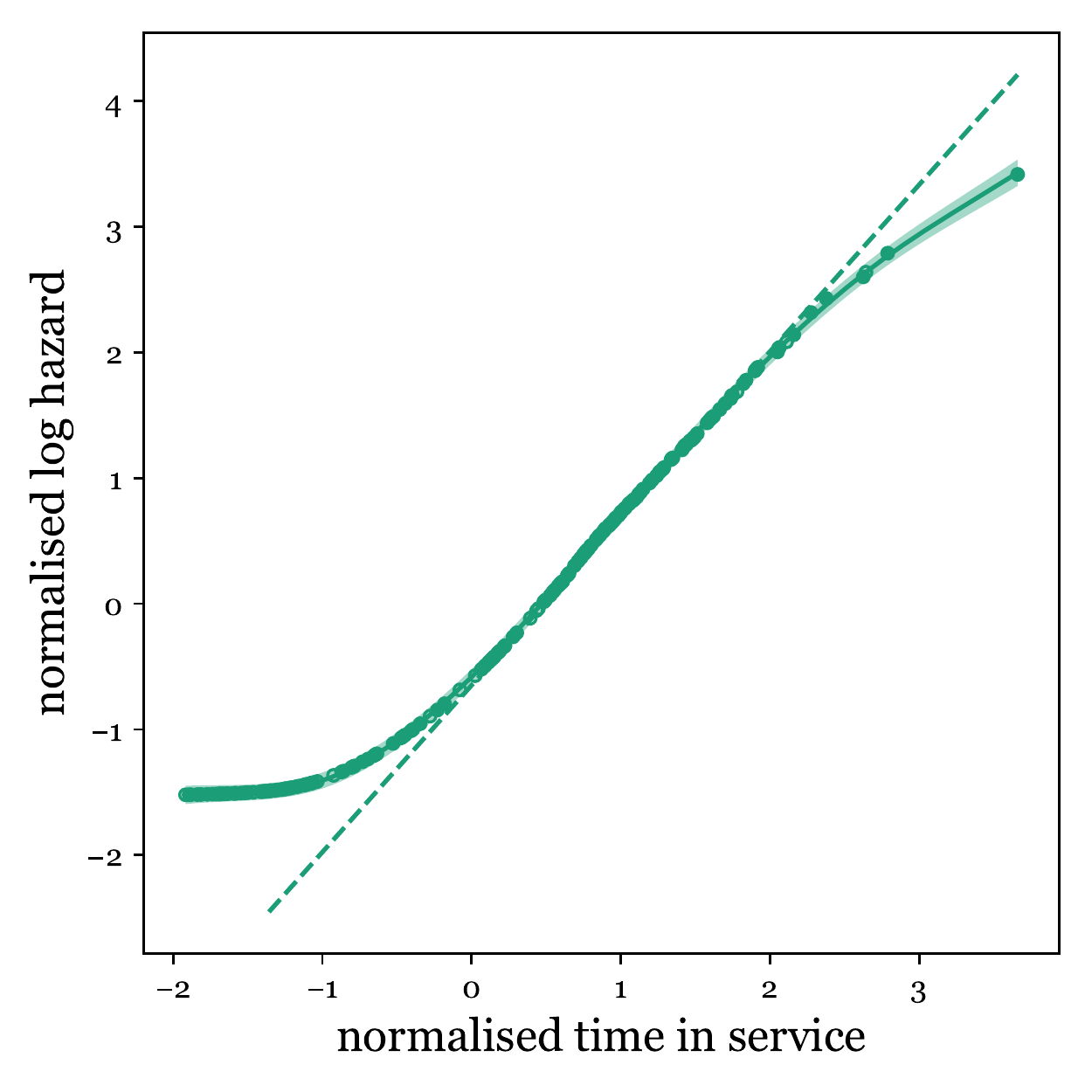}
    \caption{Basis function model for the data-rich domain {($k=1$)}. The parametric Gompertz component (\ref{e:gm-hazard}) is the dashed line and the posterior mean of the semi-parametric model (\ref{e:basis-funcs}), including splines, is the solid line.} 
    \label{fig:rich-data}
\end{figure}

To summarise, without any data, the prior postulates that the underlying log-hazard is expected to be linear, corresponding to a Gompertz survival model (\ref{e:gm-hazard}). %
The discrepancy between this simple (parametrised) behaviour and the data will be modelled by nonparametric splines, resulting in a semi-parametric regression (\ref{e:basis-funcs}) for each task. %
\Cref{fig:rich-data} visualises the implications of the model and prior, which shows the posterior predictive distribution inferred from the most data-rich domain only ($k=1$, single-task learning). %
This experiment is used to validate an appropriate number of splines for the population model, which is found to be $H=5$ through 20-fold cross-validation, presented in \ref{a:cv-scania}. %
It is intuitive to note, the same independence can, in effect, be achieved for parameters with hierarchical priors (i.e.\ $\boldsymbol{\alpha}_k$) by letting the variance of their generating distribution become very large \cite{gardner2020machine} (i.e.\ $\boldsymbol{\sigma}_\alpha \rightarrow \infty$).

Following \Cref{s:inference}, and collecting all task parameters $\mathbf{\Theta} = \{[\boldsymbol{\alpha}_k], \boldsymbol{\beta}, \boldsymbol{\mu}_\alpha, \boldsymbol{\sigma}_\alpha, \sigma \}$, the posterior distribution can be written,
\begin{align}
    p(\mathbf{\Theta}| [\mathbf{y}_k]) &= \frac{p([\mathbf{y}_k], \mathbf{\Theta})}{p([\mathbf{y}_k])} \nonumber \\
    &= \frac{p([\mathbf{y}_k] \mid \mathbf{\Theta})p(\mathbf{\Theta})}{\int p([\mathbf{y}_k], \mathbf{\Theta}) d \mathbf{\Theta}} \label{e:post-1}
\end{align}
where $p([\mathbf{y}_k] \mid \mathbf{\Theta})$ is indexed by (\ref{eq:lik-1}) and the joint prior $p(\mathbf{\Theta})$ is defined by (\ref{e:prior-expect-alpha}) to (\ref{e:prior-beta}). %
MCMC is used for inference since (\ref{e:post-1}) is intractable. %

Having conditioned on the training data $[\mathbf{y}_k]$, predictions can be made for the unobserved response $\mathbf{y}^*_k$ at $\mathbf{x}^*_k$ using the posterior predictive distribution, %

\begin{equation}
    p(\mathbf{y}_k^{*} \mid \mathbf{x}^*_k, [\mathbf{y}_k]) =  \int{p(\mathbf{y}_k^{*} \mid \mathbf{x}^*_k, \mathbf{\Theta})} p(\mathbf{\Theta} \mid [\mathbf{y}_k]) d \mathbf{\Theta}    
    \label{eq:post_pred1}
\end{equation}

(Conditioning on $\mathbf{x}^*_k$ is included here to emphasise prediction.)

\subsection{Results}
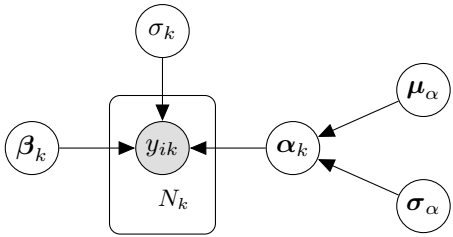
\begin{figure}[ht]
  \centering
  \begin{tikzpicture}
    \node[obs] (y) {$y_{ik}$};%
    \node[latent,right=of y] (alpha) {$\boldsymbol{\alpha}_k$}; %
    \node[latent,left=of y] (beta) {$\boldsymbol{\beta}_k$}; %
    \node[latent,above=of y, yshift=-1ex] (sigma) {$\sigma_k$}; %
    \node[latent,right=of alpha,yshift=2em] (am) {$\boldsymbol{\mu}_{\alpha}$};
    \node[latent,right=of alpha,yshift=-2em] (as) {$\boldsymbol{\sigma}_{\alpha}$};
    \plate [inner sep=2ex] {plate1} {(y)} {$N_k$}; %
    \edge {sigma,alpha,beta} {y}
    \edge {am,as} {alpha}
    \end{tikzpicture}
    \caption{DGM for independent linear models.
    }\label{f:indep-mdls}
\end{figure}

To motivate sharing information within the fleet, the regression tasks for each sub-fleet are initially learnt independently. %
This corresponds to learning separate (task-specific) parameters, which are independent, preventing the flow of information via correlated variables or tied parameters. %
The separated models can be visualised by removing the $K$ plate from the DGM in \Cref{f:hb-me-lr}, while including $k$-subscripts for $\sigma^2$ and $\boldsymbol{\beta}$. %
\Cref{f:indep-mdls} presents these updates. %

\begin{figure*}[t]
    \centering\includegraphics[width=.9\linewidth]{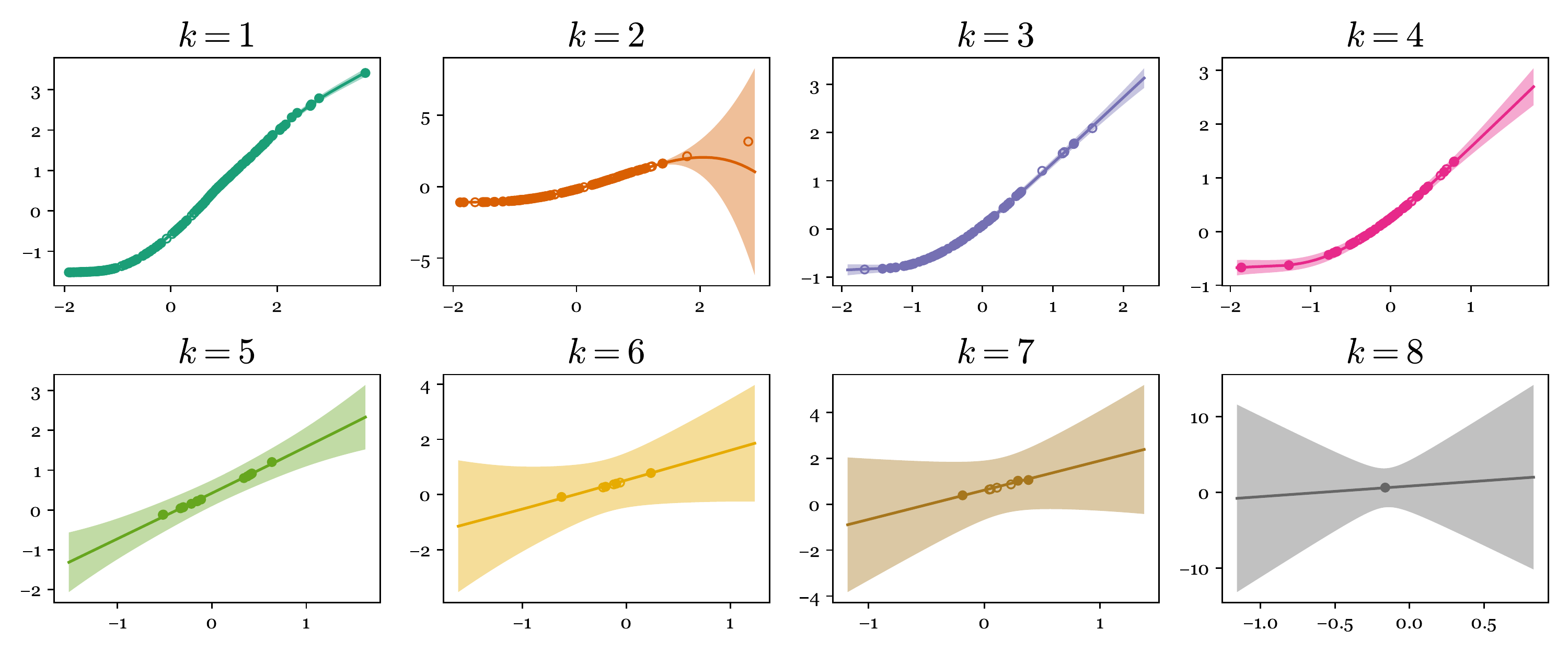}
    \caption{Posterior predictive distribution $p(\mathbf{y}^*_k | \mathbf{x}_k^*, \mathbf{x}_k, \mathbf{y}_k)$: the mean and three-sigma deviation for $K$ independent regression models.}
    \label{f:no-pooling}
\end{figure*}

\Cref{f:no-pooling} shows the resulting domain-wise regression (i.e.\ single task learning). %
The posterior-predictive distributions $p(\mathbf{y}^*_k | \mathbf{x}_k^*, \mathbf{x}_k, \mathbf{y}_k)$ make sense under the model/prior formulation, however, independent models fail to consider that valuable information might be shared between the task relationships. %
In turn, the posterior predictive distribution presents large uncertainty, especially in sparse domains. %

\begin{figure}[ht]
    \centering\includegraphics[width=.9\linewidth]{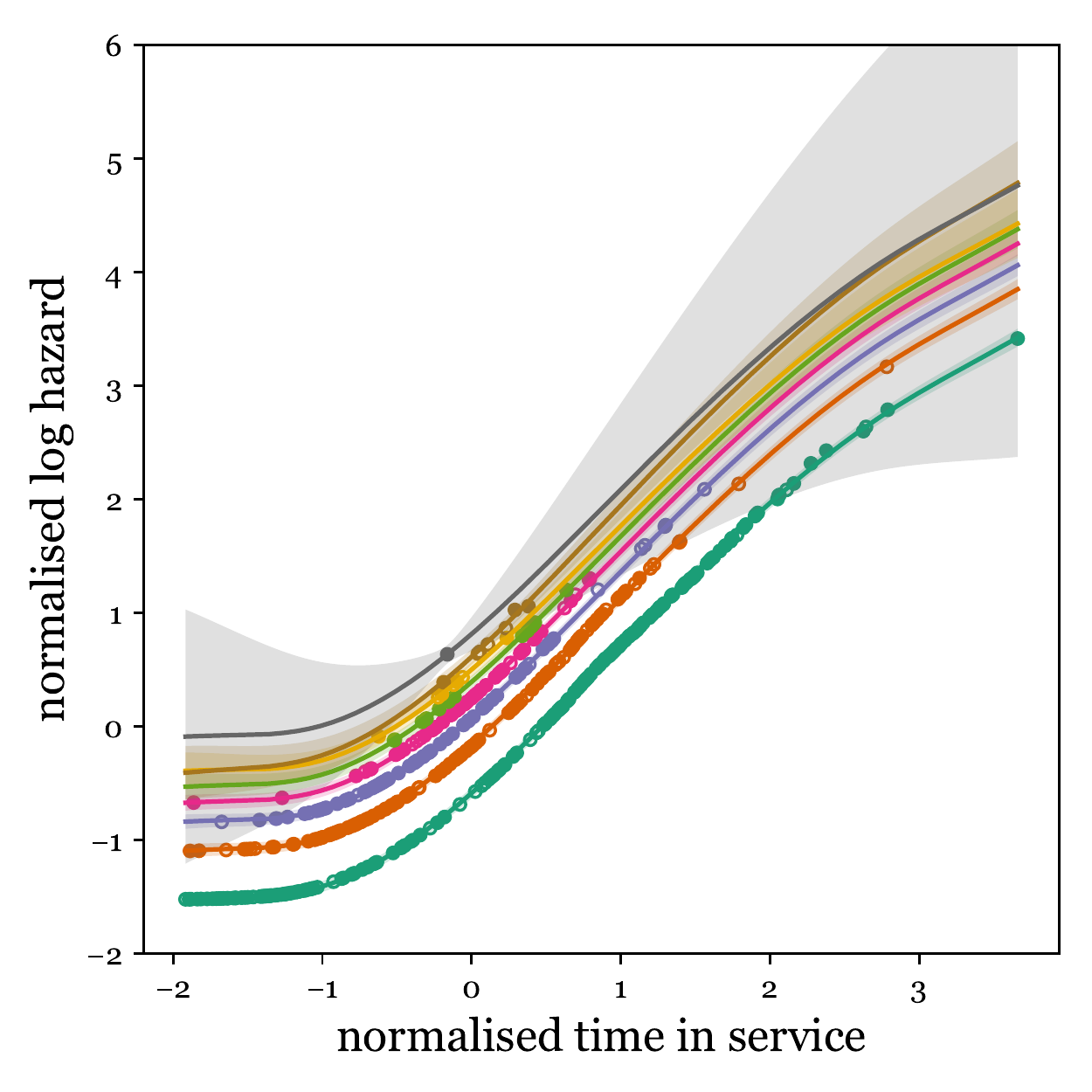}
    \caption{Posterior predictive distribution $p(\mathbf{y}^*_k | \mathbf{x}_k^*, \{\mathbf{x}_k, \mathbf{y}_k\}_{k=1}^K)$: the mean and three-sigma deviation for multitask learning with mixed effects.} %
    \label{f:functions-MEM}
\end{figure}

Hierarchical modelling is now utilised to learn the parameters in a combined inference from the population data. %
The mean and standard deviation of samples drawn from the multi-task learning posterior predictive distribution are shown in \Cref{f:functions-MEM}. %
Visually, the predictive distributions $p(\mathbf{y}^*_k | \mathbf{x}^*_k, \left\{\mathbf{x}_k, \mathbf{y}_k\right\}_{k=1}^K)$ better represent belief of the underlying task functions by leveraging information between domains. %
In particular, information from data-rich domains ($k \in \{1,2,3,4\}$) informs the (fixed) nonlinear effect. %

The predictive (log) likelihood for out-of-sample test data (25\%) is evaluated for a large number of trials (100) via bootstrap sampling \cite{murphy2012machine}. %
The combined population log-likelihood $\mathcal{L}$ increases significantly, from 355 to 410, highlighting improvements following inference at the fleet level. %
\Cref{t:alt-pll} presents the relative changes for each task, %
where STL is single-task learning, MTL is multi-task learning, CP is complete pooling, and CRL is CORAL for joint adaptation\footnote{Domain $k=8$ is excluded since there is only one observation in the historical fleet data.}. %
Compared to STL there is a relative improvement in all domains (other than $k=7$) especially those domains with sparse training data. %
In particular, leveraging information enables more reliable extrapolation to late hours in service where the test data are likely to be sparse. %
It is believed the likelihood decrease occurs in domain $k=7$, since the sub-fleet labelling may be unreliable -- the hazard data could in fact represent more than one group when observing \Cref{fig:truck-data}. Improvements to the labelling procedure are discussed as future work, \Cref{s:conc}.

Complete pooling (CP) and Correlation alignment (CRL) benchmarks behave as expected. %
CP presents the lowest overall log-likelihood $\mathcal{L}$, which makes sense considering the disparity between tasks. %
CRL successfully improves from CP by transforming the source data (all remaining domains) into the target $k$, especially when $k=7$. %
However, the total likelihood remains lower than STL, which indicates a high risk of negative transfer -- in fact, CRL improves predictions in only $k = \{6,7\}$.

\begin{table}[h]
    \centering
    \caption{Out-of-sample (average) predictive log-likelihood for 25\% test data: $\log p(\mathbf{y}^*_k | \mathbf{x}^*_k)$}
    \resizebox{\linewidth}{!}{%
    \begin{tabular}{| c || c | c | c | c | c | c | c || c |}
    \hline
     model & $k=1$ & $k=2$ & $k=3$ & $k=4$ & $k=5$ & $k=6$ & $k=7$ & $\mathcal{L}$\\
     \hline
        CP & -24.13 & 0.84 & -4.49 & -4.02 & -2.41 & -2.77 & -6.50  & -43.49 \\
        CRL & 79.29 & 49.50 & 20.07 & 11.38 & 4.24 & 2.94 & \textbf{5.47} & 172.88 \\
      STL & 150.24 & 94.24 & 57.66 & 47.04 & 8.51 & -3.17 & {0.95}  & 355.47\\
      \textit{MTL} & \textbf{166.18} & \textbf{98.23} & \textbf{64.78} & \textbf{58.09} & \textbf{25.7} & \textbf{10.17} & -13.58  & \textbf{409.57} \\
      \hline
    \end{tabular}}
    \label{t:alt-pll}
\end{table}

\begin{figure*}[t]
    \centering\includegraphics[width=.9\linewidth]{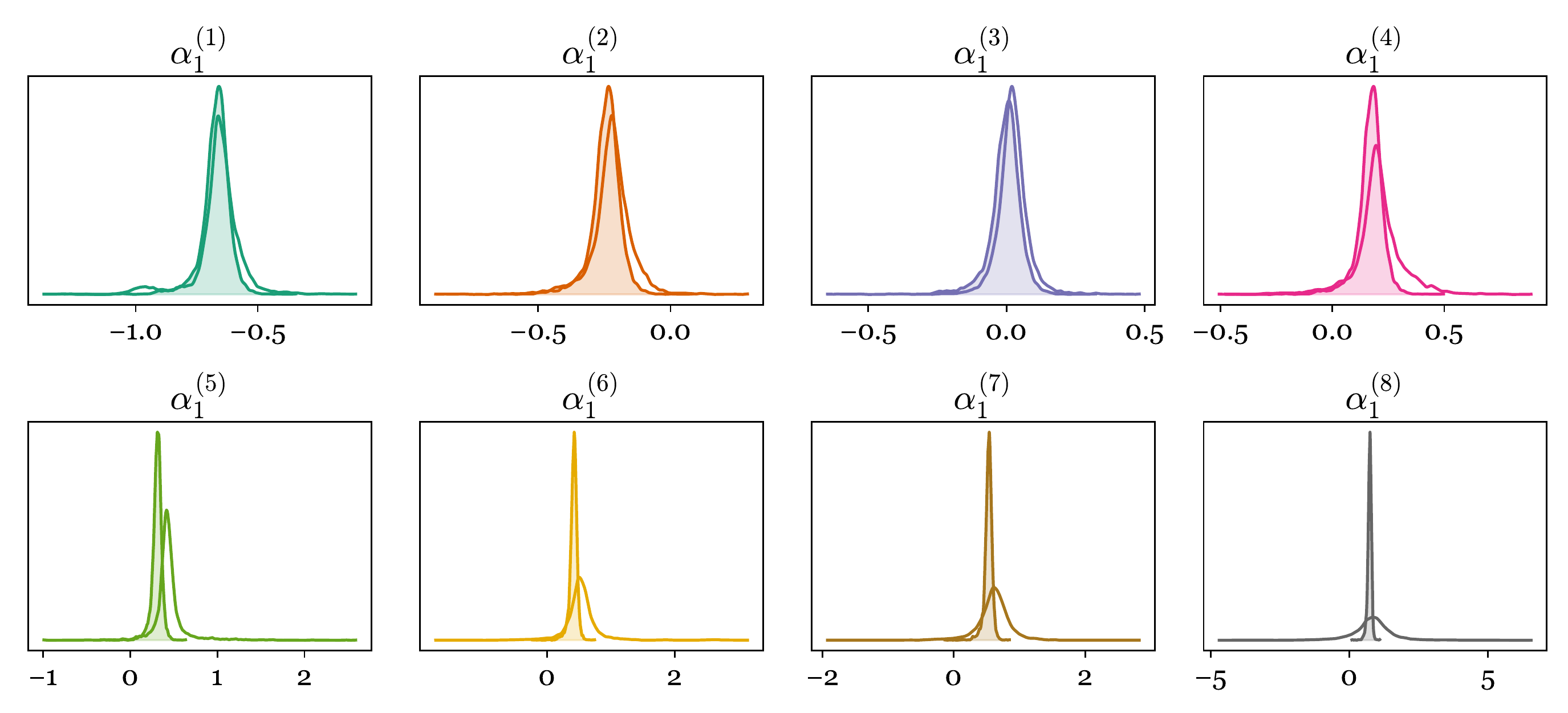}
    \caption{Variance reduction in the posterior distribution of the intercept parameters $\alpha_1^{(k)}$ for alternator components. Independent models (hollow) compared to population-level modelling (shaded).}
    \label{f:posts-alpha1}
\end{figure*}

\begin{figure*}[t]
    \centering\includegraphics[width=.9\linewidth]{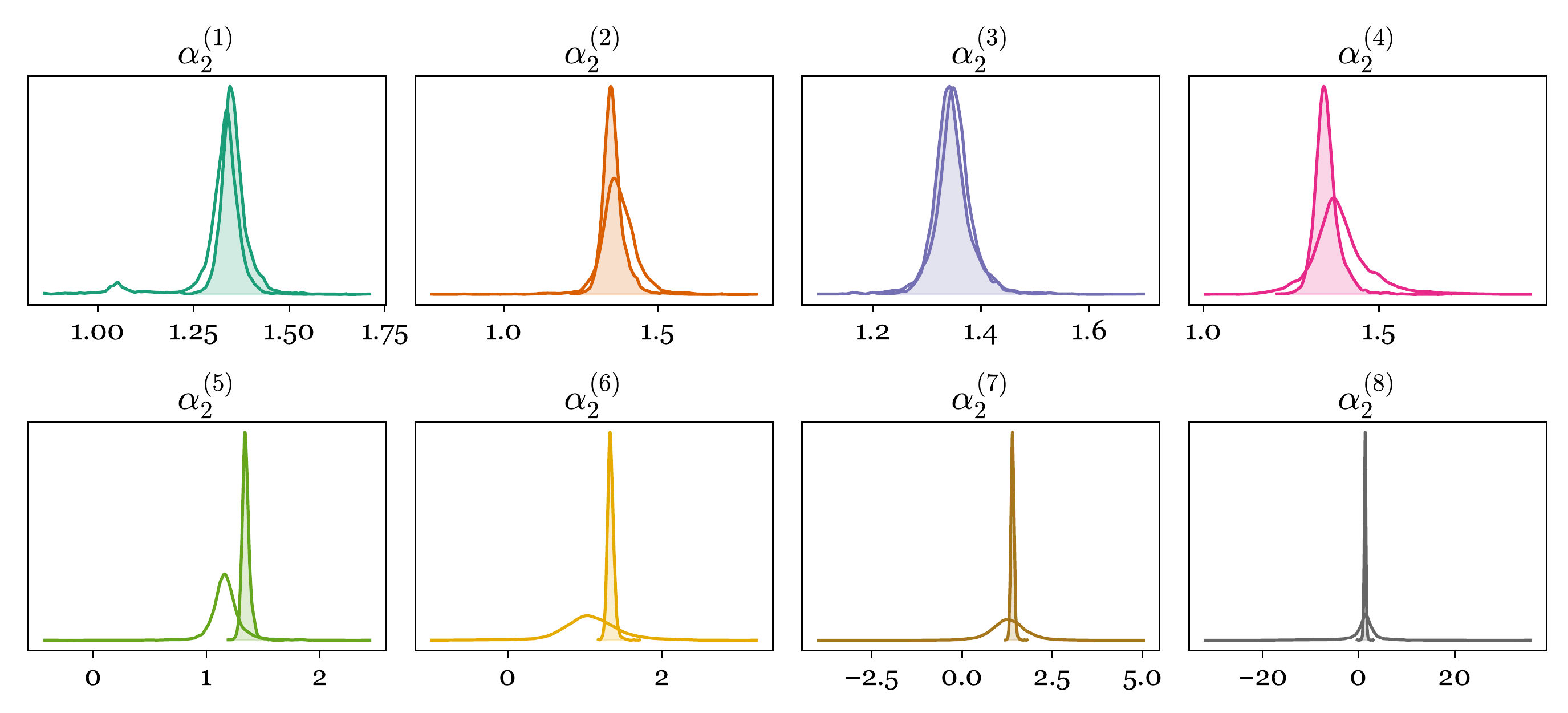}
    \caption{Variance reduction in the posterior distribution of the slope parameters $\alpha_2^{(k)}$ for alternator components. Independent models (hollow) compared to population-level modelling (shaded).}
    \label{f:posts-alpha2}
\end{figure*}

Reductions in the posterior variance of the parameters via multi-task learning are also considered, compared to single-task learning. %
\Cref{f:posts-alpha1,f:posts-alpha2} show the posterior distribution of the slope and intercepts respectively: %
these parameters correspond to the random (linear) effect of the Gompertz model $\alpha_1 + \alpha_2 t$ (\ref{e:gm-hazard}). %
Variance reductions are most significant in sparse domains (bottom row) and less significant in the data-rich domains (top row). %
This follows intuition since the population model allows sparse domains to \textit{borrow} information via the shared parent nodes $\{\boldsymbol{\mu}_\alpha, \boldsymbol{\sigma}^2_\alpha\}$ while the data-rich domains are largely unaffected. %
Quantitatively, the average reduction in standard deviation for the (interpretable) linear weights is 90\% and 73\% for the slopes and intercepts respectively. %

\begin{figure}[ht]
    \centering\includegraphics[width=.7\linewidth]{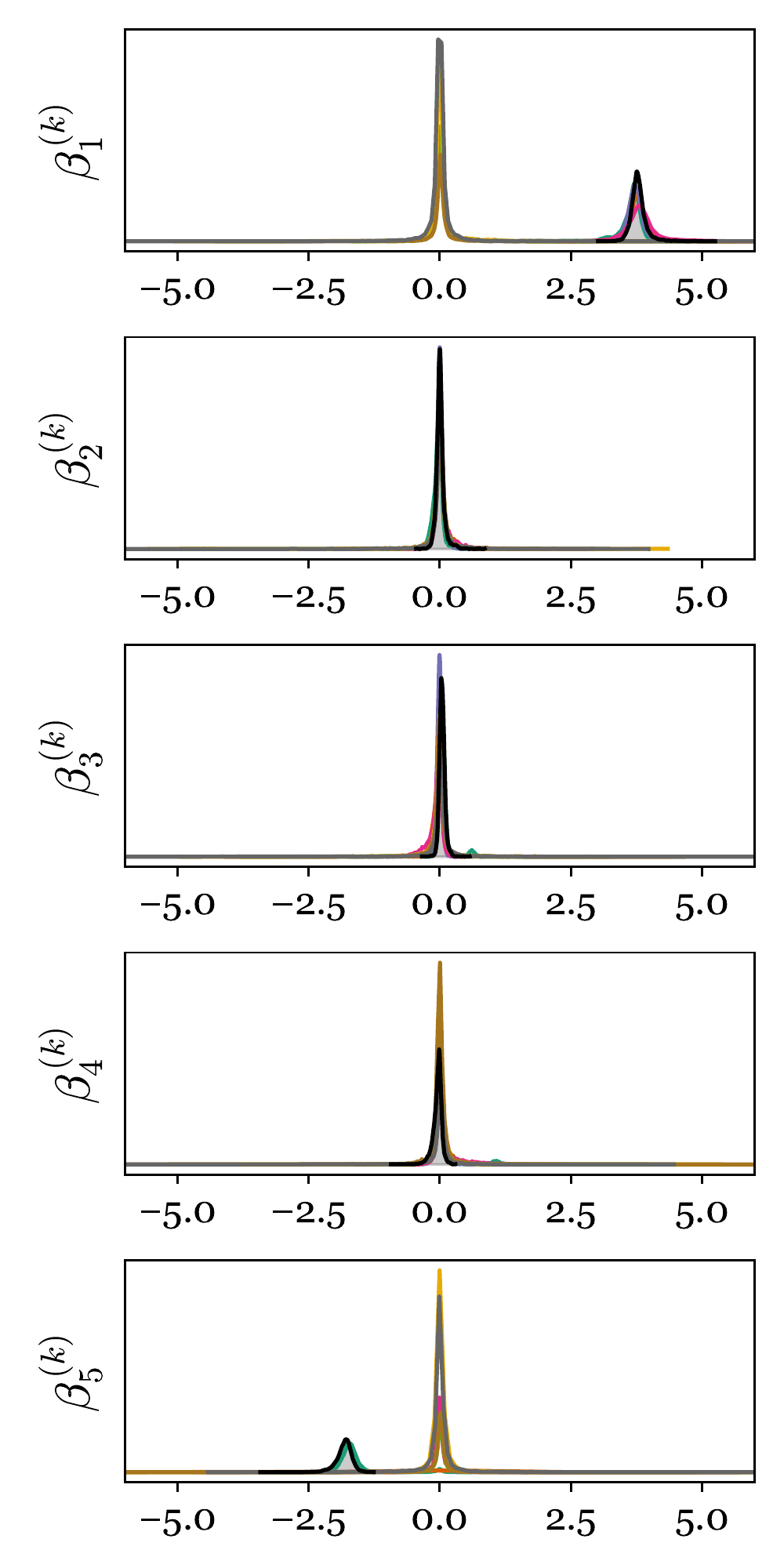}%
    \caption{Posterior distribution of the weight parameters $\boldsymbol{\beta}_{(k)} = \{\beta_h^{(k)}\}_{h=1}^H$. Comparison between the tied population-level parameters (grey shaded) and independent models (hollow) for each domain $k \in \{1, \ldots, 8\}$. Zoomed sections for $h = 1$ and $5$ are provided in \ref{a:zoomed}.}
    \label{f:posts-beta}
\end{figure}

\Cref{f:posts-beta} shows the posterior distribution of the fixed weights $\boldsymbol{\beta}$. %
Under the prior specification, these weights adaptively deviate from zero to model the discrepancy from the linear effect in sparse/compact regions of the input (via nonparametric splines). %
Building on intuition, by tying these parameters, the expected values shift towards the expectation of the data-rich, independent models ($k \in \{1,2,3,4\}$). %
In other words, in the population-level inference, the fixed effect is learnt from the domains which have data to describe it. %

\begin{figure}[ht]
    \centering\includegraphics[width=.8\linewidth]{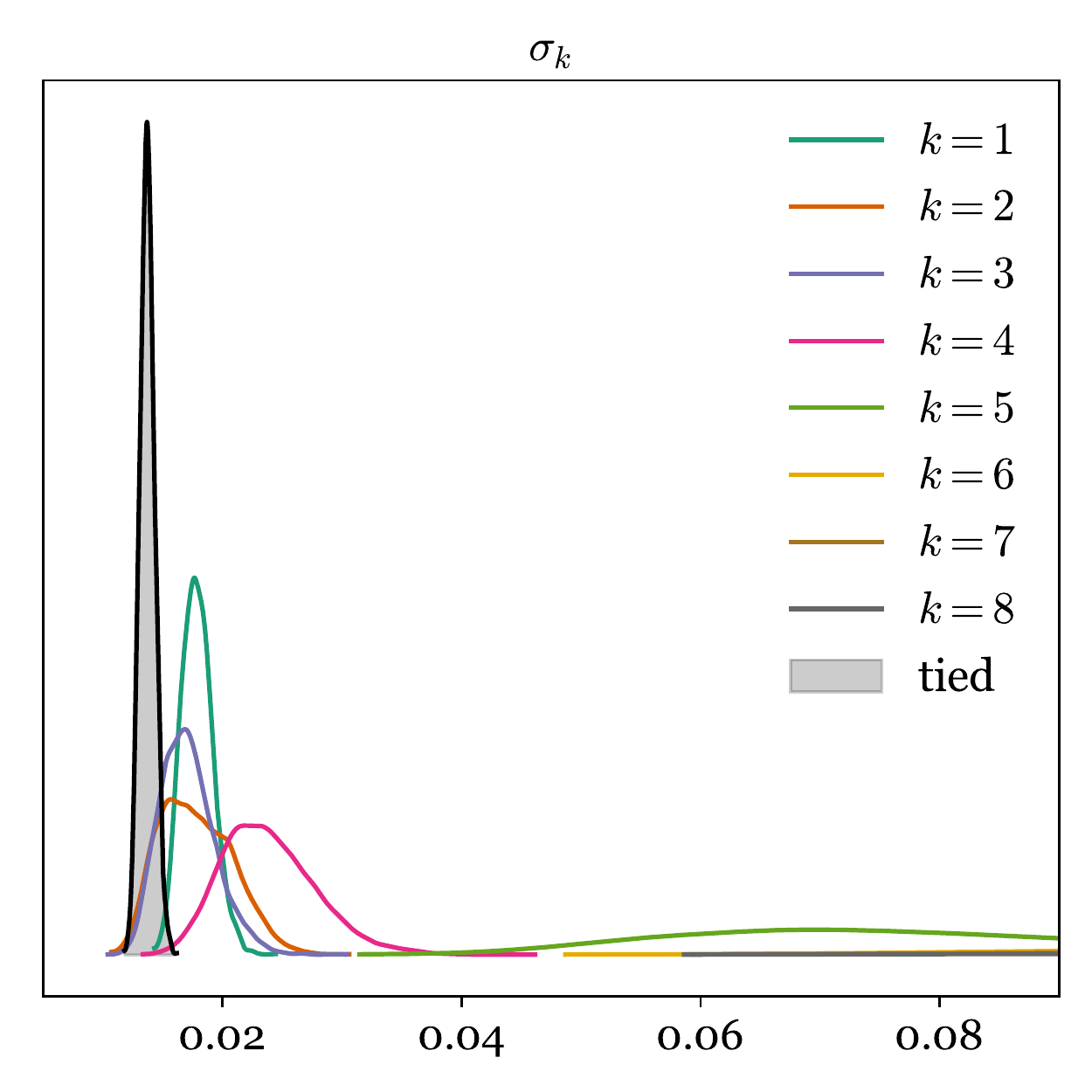}
    \caption{Posterior distribution of the noise parameter $\sigma_{(k)}$. Comparison between the tied population-level parameters (grey shaded) and independent models (hollow) for each domain $k \in \{1, \ldots, 8\}$.}
    \label{f:posts-sigma}
\end{figure}

Likewise, \Cref{f:posts-sigma} shows improvements in the estimate of $\sigma_{(k)}$ when tying the noise effect. %
The posterior variance is reduced, while the expected values indicate a lower noise variance. %
This should be expected since by pooling the data to learn $\sigma_{(k)}$ the training set is effectively extended; in turn, the posterior moves further away from the weakly informative prior (\ref{e:prior_simga}).

\subsection{Modelling additional failures and the risk of negative transfer}
The assumptions which select the tied parameters are critical -- this caveat is widely acknowledged. %
If any assumptions prove inappropriate or non-general, the multi-task learner can risk negative transfer, whereby predictions are worse than conventional (i.e.\ single-task) learning -- i.e.\ in this case, independent models. %
In a probabilistic setting, negative transfer manifests as inappropriate inter-task correlations; to control these dependencies one could utilise shrinkage \cite{gelman2006data} or automatic relevance determination \cite{tipping2001sparse} (between tasks) to protect against such issues; these ideas are suggested for future work. %

\begin{figure}[t]
    \centering\includegraphics[width=.9\linewidth]{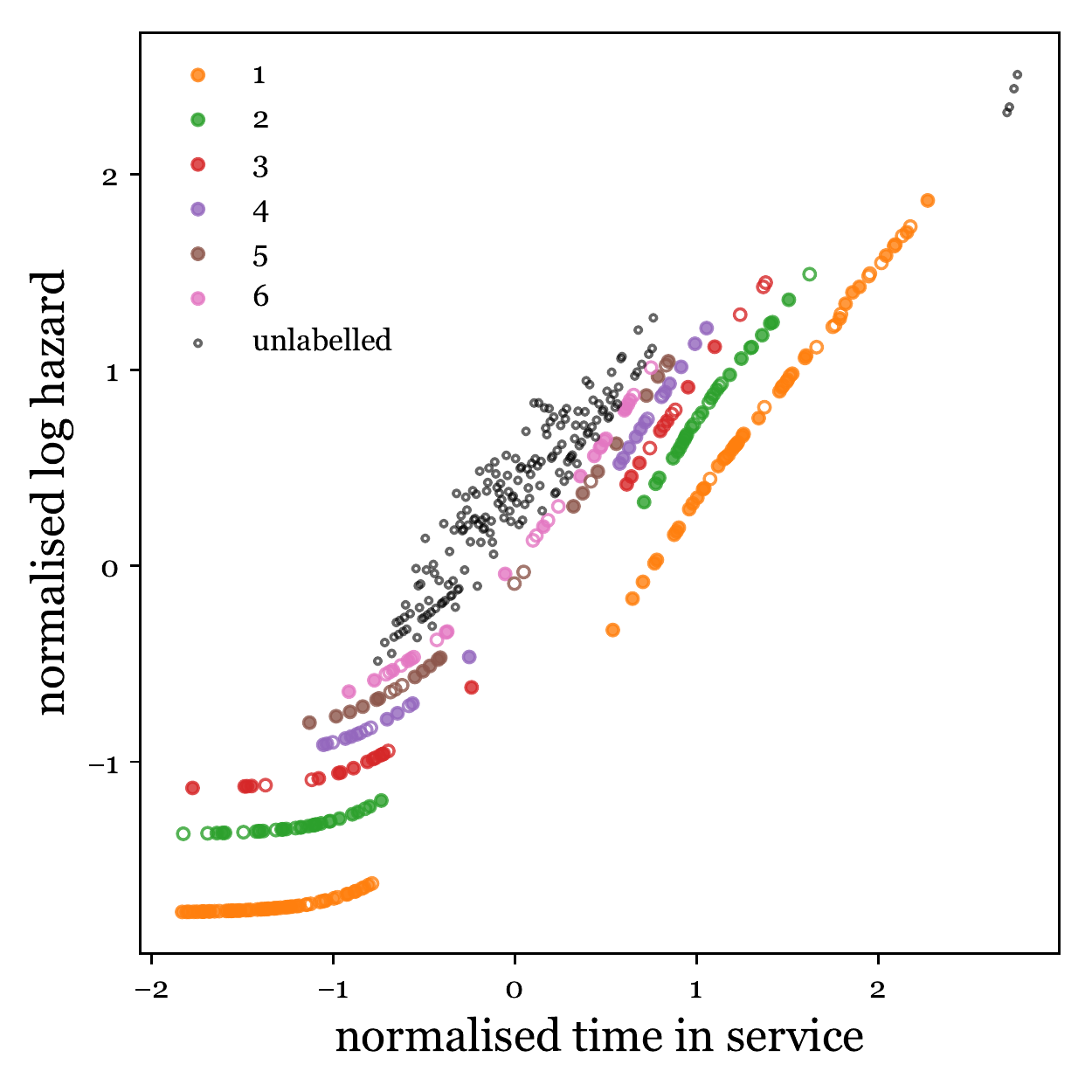}
    \caption{Log hazard data for turbochargers in the truck fleet. Training and testing markers are $\bullet$ and $\circ$ respectively. Colours correspond to sub-fleet labels (\textit{\ref{a:post-pub} post-publication note}.)}
    \label{fig:truck-data-TCs}
\end{figure}

To highlight concerns of negative transfer, the empirical hazard data are considered from another component in the same fleet of vehicles, turbochargers. %
The survival data are presented \Cref{fig:truck-data-TCs}, which are calculated following the same procedure as the alternators. %
Critically, manually labelling the alternator data is problematic, %
since it becomes infeasible to categorise observations as the generating functions become more compact, or towards the end of the operational life. %
The associated \textit{unlabelled} data are highlighted with small $\circ$ markers in \Cref{fig:truck-data-TCs} (\textit{\ref{a:post-pub} post-publication note}). %

There are various options when considering these data. %
One could treat the observations as a single (pooled) sub-fleet or task, with a large expected variance; alternatively, the labels themselves could be treated as an additional latent variable, such that categorisation into task groups is unsupervised. %
Here, the unlabelled data are removed during preprocessing, since modelling them is out of the scope of this work; alternative solutions are proposed in the concluding remarks, \Cref{s:conc}. %
The resulting turbocharger dataset has 287 (normalised) observations over six tasks, such that $k \in \{1, 2, \ldots, 6\}$, and the number of observations in each domain is as follows,

\begin{center}
\begin{tabular}{| c | c | c | c | c | c || c |}
\hline
 $N_1$ & $N_2$ & $N_3$ & $N_4$ & $N_5$ & $N_6$ & $\sum_{k=1}^K N_k$\\
 \hline
 112 & 60 & 32 & 28 & 25 & 30 & 287 \\
 \hline
\end{tabular}
\end{center}

As before, the data are split into 75\% training and 25\% test sets. %

From \Cref{fig:truck-data-TCs}, one observes that the turbocharger hazard data are similar to \Cref{fig:truck-data} (alternators). %
Since the components operate within the same fleet of vehicles, it is assumed that information can be shared between the associated predictors by extending the task-set in the hierarchical model. %
A na\"ive approach assumes the same formulation of mixed effects, %
and simply extends the total number of tasks such that ${K = 14}$ (i.e.\ $8 + 6$) then infers the parameters from both alternator and turbocharger hazard data. %
\ref{a:tc-initial} presents the posterior predictive distribution of such a model. %
While the model interpolates well, the extrapolation behaviour\footnote{At the population level, this is not extrapolation, since the response at late hours in service is learnt from the alternator domain.} is problematic for later hours in service. %
This is because the model assumes that the discrepancy (from the Gompertz model) is equivalent for both components, as the nonparametric weights $\boldsymbol{\beta}$ remain tied over all tasks. %
The unlabelled data are evidence that this assumption is inappropriate, as the model would generalise poorly to these data, plotted in \ref{a:tc-initial}. %
The resultant model would have a high risk of negative transfer. %

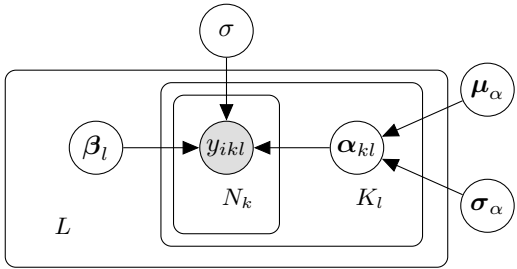
\begin{figure}[ht]
  \centering
  \begin{tikzpicture}
    \node[obs] (y) {$y_{ikl}$};%
    \node[latent,right=of y] (alpha) {$\boldsymbol{\alpha}_{kl}$}; %
    \node[latent,left=of y] (beta) {$\boldsymbol{\beta}_l$}; %
    \node[latent,above=of y, yshift=-1ex] (sigma) {$\sigma$}; %
    \node[latent,right=of alpha,yshift=2em] (am) {$\boldsymbol{\mu}_{\alpha}$};
    \node[latent,right=of alpha,yshift=-2em] (as) {$\boldsymbol{\sigma}_{\alpha}$};
    \plate [inner sep=3ex] {plate2} {(y)(alpha)} {$K_l$}; %
    \plate [inner sep=2ex] {plate1} {(y)} {$N_k$}; %
    \plate [inner sep=5ex, yshift=-1ex, label={[xshift=-13ex, yshift=-14ex]$L$}] {plate3} {(beta) (y) (alpha)} {}; 
    \edge {sigma,alpha,beta} {y}
    \edge {am,as} {alpha}
    \end{tikzpicture}
    \caption{DGM of hierarchical linear regression with mixed effects. Introducing a higher-level group, such that the total number of tasks is $L \times K_l$.
    }\label{f:mem-2}
\end{figure}

Instead, the mixed effect model is reformulated, whereby a separate, nonparametric discrepancy $\{\boldsymbol{\beta}_l\}_{l=1}^L$ is learnt for the alternator ($l=1$) and turbocharger ($l=2$) tasks -- introducing two higher-level subgroups, such that $L=2$. %
As before, the parameters of the linear component remain correlated via the shared parent nodes, allowing knowledge transfer between all 14 tasks (both alternators and turbochargers). %
In turn, the model and prior now postulate a varying underlying linear trend for all tasks (the Gompertz model); however, the discrepancy from this behaviour is component-specific (a separate $\boldsymbol{\beta}_l$ for each component). %
The modifications can be visualised by updating the DGM from \Cref{f:hb-me-lr} to include higher-level subgroups $l \in \{1, 2\}$, presented in \Cref{f:mem-2}, where \ $l=1$ alternators or $l=2$ turbochargers. %

A key difference is the new $L$-plate and the associated subscripts: %
$K_l$ is the number of sub-fleets for each component, such that $K_1 = 8$ (alternators) or $K_2=6$ (turbochargers); while $\boldsymbol{\beta}_l$ indicates a separate (independent) weight vector for each component. %
The collected tasks become,
\begin{align}
\left\{ \bigg\{ \mathbf{y}_{kl} = \underbrace{\boldsymbol{\Phi}_{kl} \boldsymbol{\alpha}_k}_{\textrm{random}} + \underbrace{\boldsymbol{\Psi}_{kl} \boldsymbol{\beta}_l}_{\textrm{fixed}} + \boldsymbol{\epsilon}_{kl} \bigg\}_{k=1}^{K_l} \right\}_{l=1}^L \label{e:mem-model-eqns2}
\end{align}

\begin{figure*}[ht]
    \includegraphics[width=.9\linewidth]{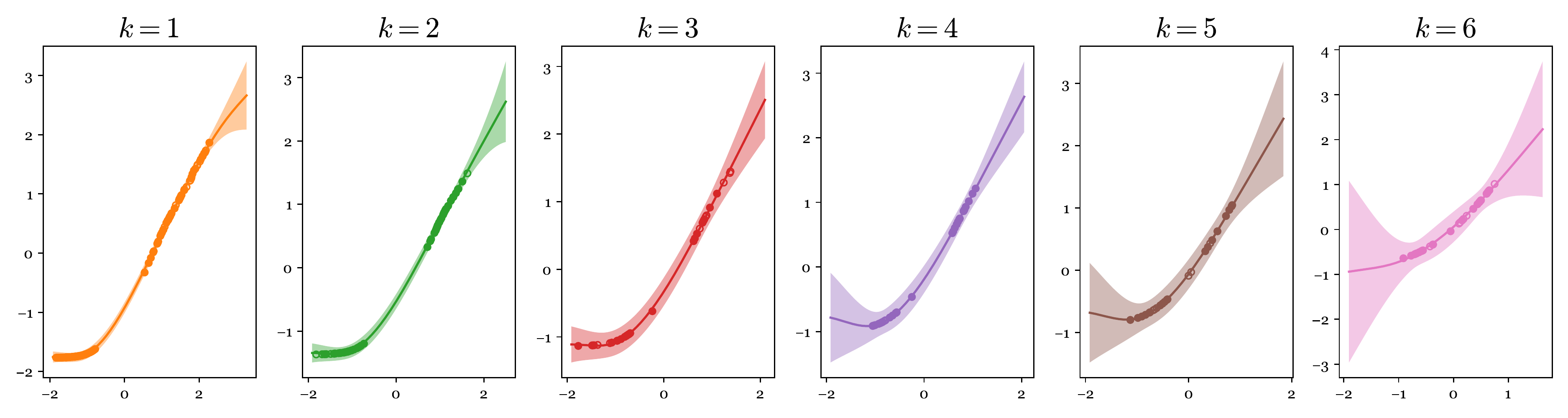}
    \centering\includegraphics[width=.7\linewidth]{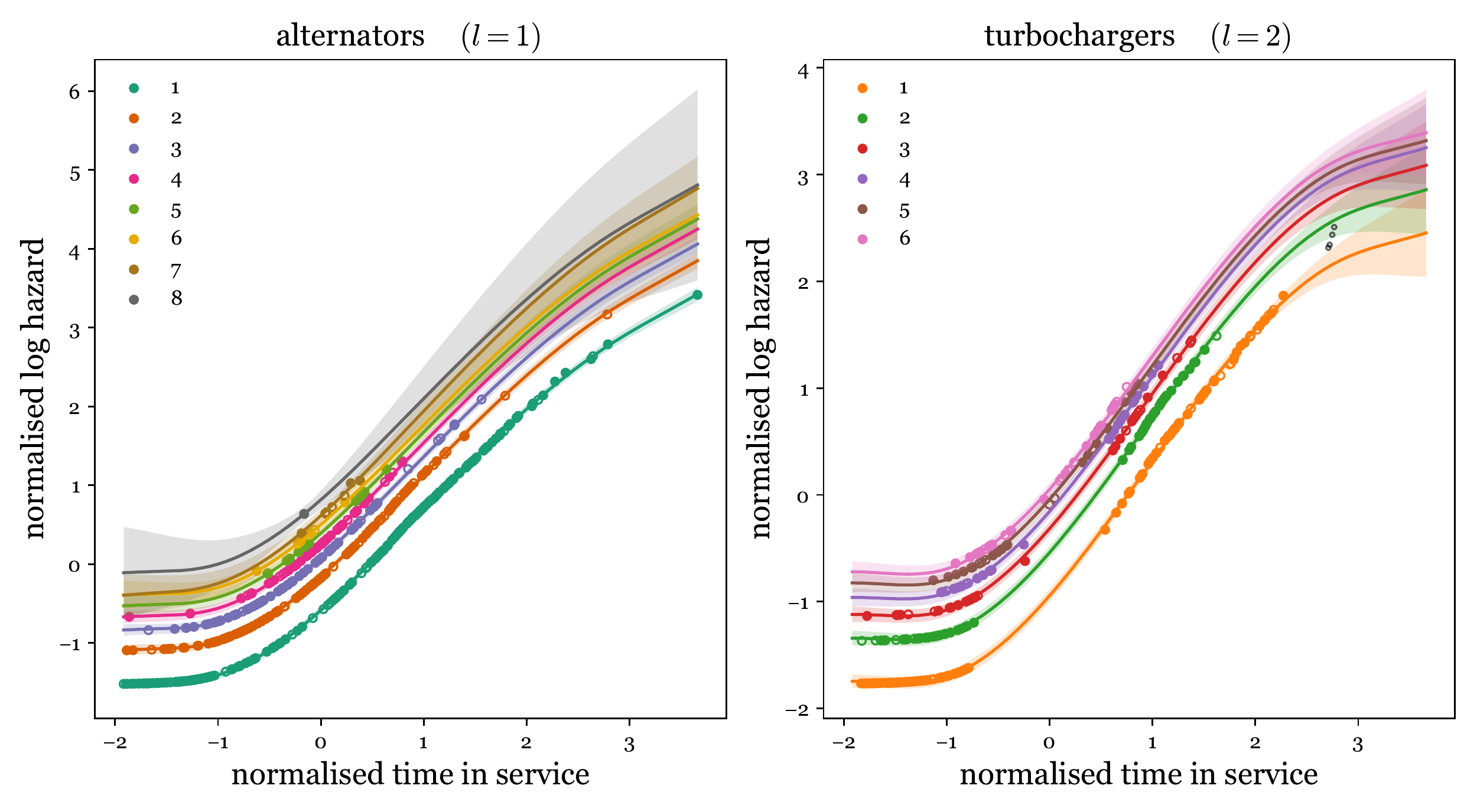}
    \caption{Posterior predictive distribution, the mean and three-sigma deviation for: (top) $K$ independent regression models of turbocharger hazard $p(\mathbf{y}^*_{kl} | \mathbf{x}_{kl}^*, \mathbf{x}_{kl}, \mathbf{y}_{kl})$. (bottom) multitask learning with mixed effects for all turbocharger and alternator tasks $p(\mathbf{y}^*_{kl} | \mathbf{x}_{kl}^*, \{\{\mathbf{x}_{kl}, \mathbf{y}_{kl}\}_{k=1}^{K_l}\}_{l=1}^L)$.} %
    \label{f:all-funcs-MEM2}
\end{figure*}

In turn, the likelihood of the response is modified,
\begin{align}
y_{ikl}|& x_{ikl}, \boldsymbol{\theta}_{kl} \sim \nonumber \\ 
&\textrm{N}\left(\alpha^{(kl)}_1 + \alpha^{(kl)}_2 x_{ikl} + \sum^H_{h=1} \beta^{(l)}_h b_h(x_{ikl}),\; \sigma^2\right) \label{eq:lik-2}
\end{align}

where $\boldsymbol{\theta}_{k,l} = \{\boldsymbol{\alpha}_{k,l}, \boldsymbol{\beta}_l, \boldsymbol{\mu}_\alpha, \boldsymbol{\sigma}_\alpha, \sigma \}$ is the parameter set indexed to group $k$ and component $l$. %
\Cref{f:all-funcs-MEM2} plots the mean and standard deviation of samples drawn from the posterior distribution of the extended population model (compared to independent turbocharger models). %
By specifying component-specific weights $\boldsymbol{\beta}_l$ the representation of uncertainty improves when extrapolating in the turbocharger domain. %
Reductions in the posterior predictive distribution are also observed $p(\mathbf{y}^*_{kl} | \mathbf{x}^*_{kl})$ (ignoring other conditionals) for alternator tasks ($l=1$) since the population data have been extended for the linear component. %
Likewise, variance reductions are observed in the posterior distributions of the intercept and slope, %
visualised in \ref{a:tc-VR}. %
Quantitatively, the average reduction in standard deviation for the (interpretable) linear weights is 51\% and 67\% for the (turbocharger) slopes and intercepts respectively. %

Fleet-level inference improves the (bootstrapped) predictive log-likelihood from 570 to 646, compared to single-task learning (STL), highlighting improvements in predictive capability for the combined fleet over both components. %
The task-wise predictive likelihood is presented in \Cref{t:alt-tc-pll} for the alternator ($l=1$) and turbocharger ($l=2$) domains, compared to the same benchmarks. %
Note, however, that the likelihood fails to increase from STL for certain alternator tasks ($k = 1$ or $5$) reiterating the risk of negative transfer in the extended model. %
Ideally, the dataset should be much larger to determine if negative transfer has occurred and whether the current assumptions are appropriate. %
As before, while correlation alignment (CRL) improves on complete pooling (CP) the adaption approach is not suitable for the task set, and predictions remain worse than STL. %
The sparsity of measurements prohibits reliable transformations of the source data into the target domain. %

\begin{table}[ht]
    \centering
    \caption{Out-of-sample (average) predictive log-likelihood for 25\% test data, $\log p(\mathbf{y}^*_{kl} | \mathbf{x}^*_{kl})$. Here $l$ corresponds to the component label (alternator $l=1$ or turbocharger $l=2$) while $k$ is the sub-fleet label for each component. (The complete log-likelihood considers all groups and components $\mathcal{L}$.)}
    
    \small{Alternators: $l=1$}
    
    \vspace{4pt}
    \resizebox{\linewidth}{!}{%
    \begin{tabular}{| c || c | c | c | c | c | c | c |}
     \hline
      model & $k=1$ & $k=2$ & $k=3$ & $k=4$ & $k=5$ & $k=6$ & $k=7$ \\
     \hline
      CP & -24.13 & 0.84 & -4.49 & -4.02 & -2.41 & -2.77 & -6.50 \\
      CRL & 79.29 & 49.50 & 20.07 & 11.38 & 4.24 & 2.94 & \textbf{5.47} \\
      STL & 150.24 & 94.24 & 57.66 & 47.04 & 8.51 & -3.17 & {0.95} \\
      \textit{MTL} & \textbf{164.54} & \textbf{96.98} & \textbf{62.79} & \textbf{57.51} & \textbf{24.7} & \textbf{11.12} & -9.13 \\
     \hline
    \end{tabular}}
    \vspace{1ex}
    
    \small{Turbochargers: $l=2$}
    
    \vspace{4pt}
    \resizebox{\linewidth}{!}{%
    \begin{tabular}{| c || c | c | c | c | c | c || c |}
     \hline
      model & $k=1$ & $k=2$ & $k=3$ & $k=4$ & $k=5$ & $k=6$ & $\mathcal{L}$ \\
     \hline
      CP & -9.85 & 0.35 & -1.16 & -0.23 & -0.46 & -4.00 & -58.83 \\
      CRL & 46.96 & 23.27 & 15.13 & 8.39 & 9.85 & 10.64 & 287.12  \\
      STL & \textbf{90.37} & 48.35 & 21.63 & 17.13 & \textbf{13.73} & 23.74 & 570.41 \\
      \textit{MTL} & 81.34 & \textbf{53.14} & \textbf{35.97} & \textbf{23.94} & 11.2 & \textbf{32.17} &  \textbf{646.28} \\
     \hline
    \end{tabular}}
    \label{t:alt-tc-pll}
\end{table}

\begin{figure*}[ht]
\begin{subfigure}[t]{.49\linewidth}
    \centering\includegraphics[width=\linewidth]{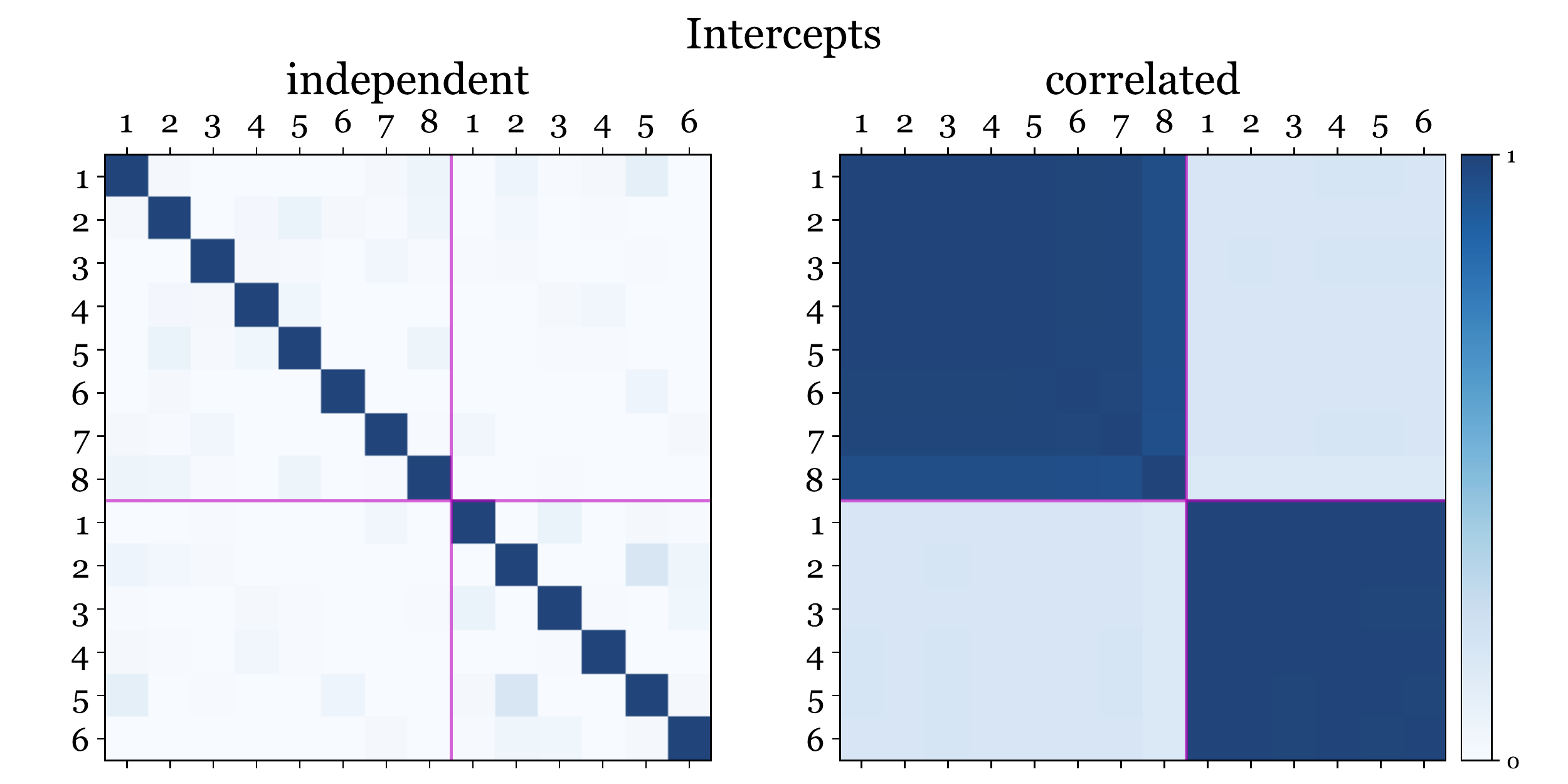}
    \caption{}\label{fig:corr-a1}
\end{subfigure}
\begin{subfigure}[t]{.49\linewidth}
    \centering\includegraphics[width=\linewidth]{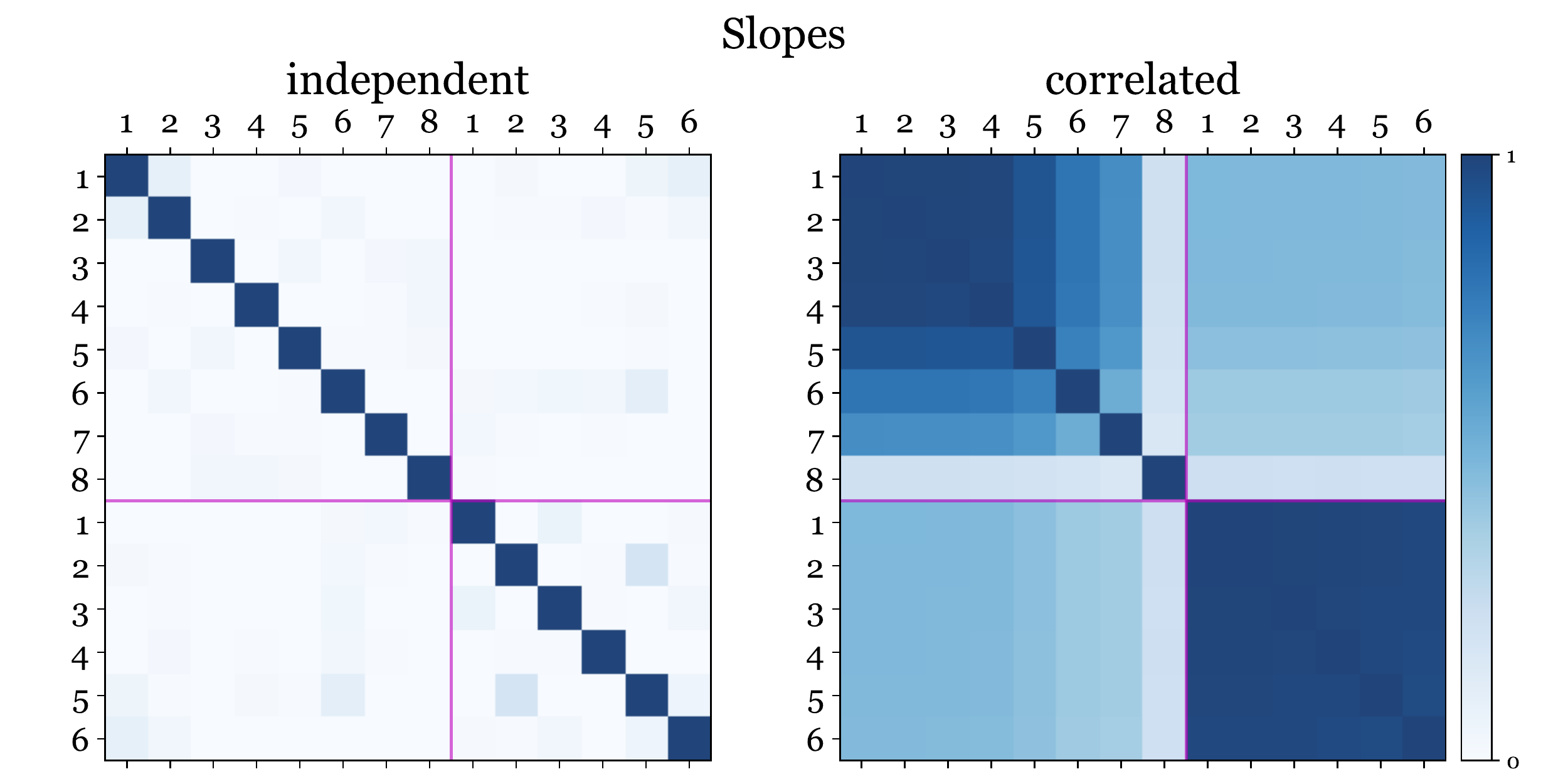}
    \caption{}\label{fig:corr-a2}
\end{subfigure}
\caption{
Pearson correlation coefficient of the conditional posterior distribution for the linear coefficients $\boldsymbol{\alpha}_k$ (slopes and intercepts). %
Purple lines separate the alternator tasks (up to 8) and turbocharger tasks (up to 6). %
}
\label{fig:corrs}
\end{figure*}

\Cref{fig:corrs} is insightful since it informs which correlations in the hierarchy \textit{transfer} or \textit{share} information between the sub-fleet~($k$) or component~($l$) groups. %
The heat-map corresponds to the Pearson correlation coefficient of the posterior distribution between variables that share parent nodes in the graphical model (i.e.\ $\boldsymbol{\alpha}_{kl}$) -- these correlations enable multi-task learning. %
Intuitively, \Cref{fig:corr-a1} shows increased correlation between the intercepts of the same component, with two clear blocks of $8 \times 8$ (alternators) and $6 \times 6$ (turbochargers). %
The intercept correlation structure is interpretable since components of the same type are likely to have a correlated baseline hazard. %

The slope correlation structure in \Cref{fig:corr-a2} is more descriptive. %
In the top left block, the alternators are less correlated as domains become more sparse (from $1 \to 8$); this makes sense since the level of correlation is reduced where there are fewer data to support task correlation. %
The effect is most obvious for $k=8$ (alternators) which only has a single training point. %
In both \Cref{fig:corr-a1,fig:corr-a2}, the structured covariance of $\boldsymbol{\alpha}_{kl}$ highlights how inter-task correlation contributes to variance reduction in the fleet-model. %

\subsection{Practical implications}

In the field, use-type labels could be used to make sub-fleet (rather than global) predictions, which has major implications when informing efficiency or safety-critical interactions with the fleet. %
For example, task-specific estimations of remaining useful life would be associated with less uncertainty, and the hierarchical model allows both population estimates (from the generating distributions) and task-specific estimates. %
These multilevel predictions present a key contribution of this work; in turn, a multilevel decision process could be designed for more reliable interactions with the fleet -- such as vehicle servicing or component replacement. %
A hypothetical decision process is demonstrated in the next case study. %

\section{Wind Farm Power Prediction}\label{s:wind-farms}

\begin{figure}[ht]
    \includegraphics[height=.66\linewidth]{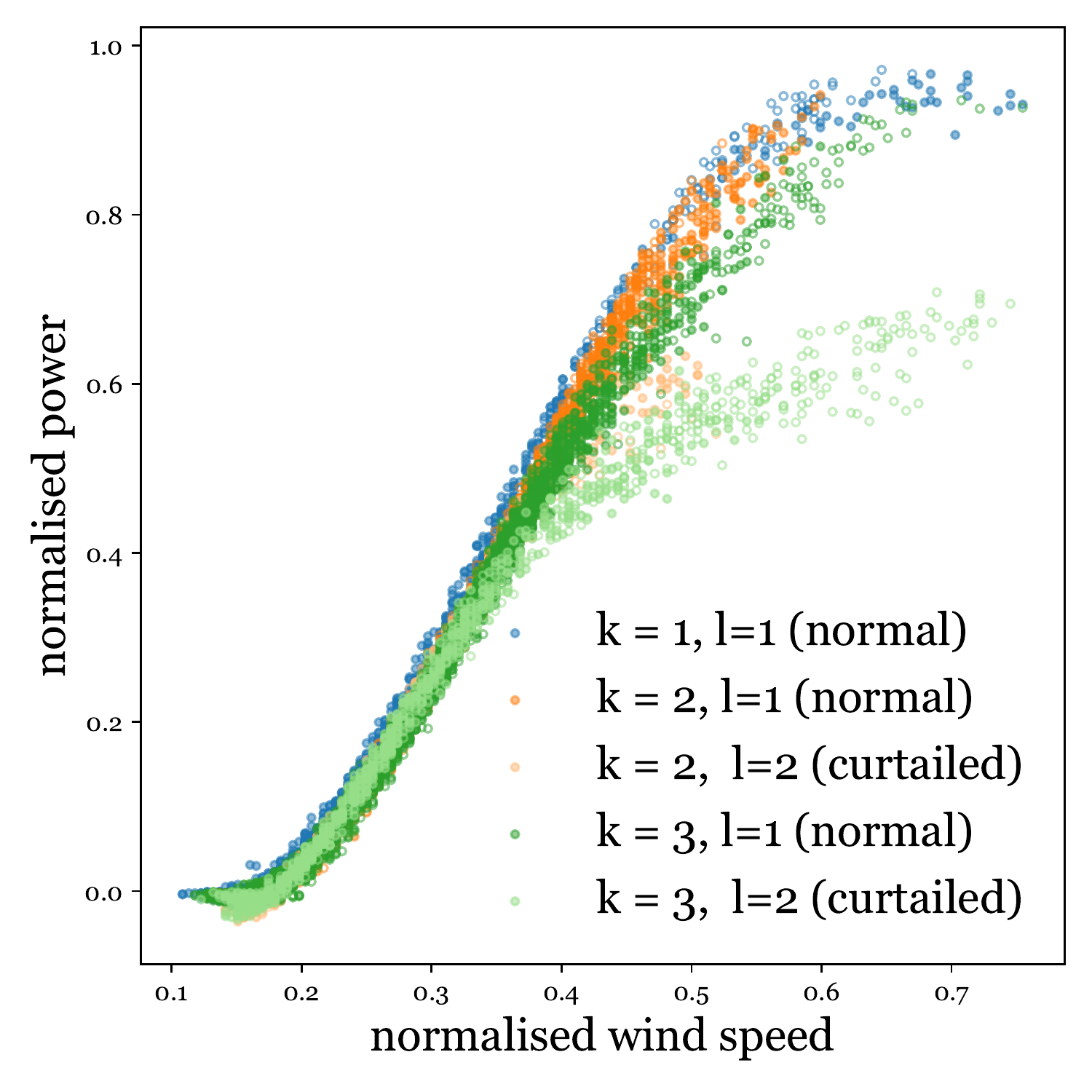}
    \centering\includegraphics[height=.66\linewidth]{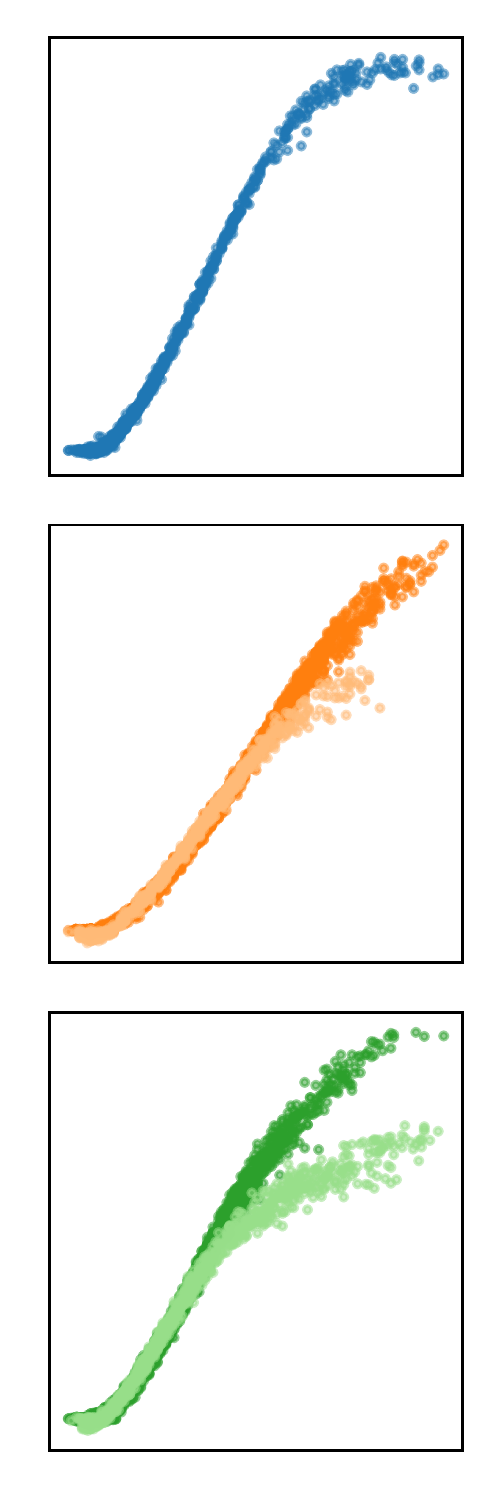}
    \caption{Power-curve data from three $k \in \{1,2,3\}$ wind turbines of the same make and model. Relationships correspond to normal $l=1$ and ideal $l=2$ operation.} %
    \label{f:PC-data}
\end{figure}

To demonstrate the wide applicability of hierarchical models, power prediction is presented for a wind farm case study. %
\Cref{f:PC-data} shows power curve data, including curtailments, provided by Visualwind and recorded from three operational turbines. %
The turbines are the same make and model but in different locations. %
As before, the data are normalised in view of data sensitivity and certain (specific) details are omitted -- the same comments regarding interpretability, data truncation, and censorship apply. %
The work in \cite{bull2021bayesian} demonstrates a suitable method to represent similar normal and curtailed functions in a combined model; however, each function $f_k$ is assumed independent -- in turn, there is no knowledge transfer between task parameters. %
Here, knowledge transfer is enabled by correlating the regression models in a hierarchical formulation. %

There are 10,581 observations in total. %
The data were labelled in weekly subsets, according to turbine $k \in \{1, 2, 3\}$ and operational condition (normal or curtailed) $l \in \{1, 2\}$. %
Each point corresponds to a 10-minute average of power $y_{ikl}$ and wind speed $x_{ikl}$. %
The first turbine has 2 weeks of data, the second has 4 weeks, and the third has 11.5 weeks. %
Missing values and very sparse outliers were removed from the dataset (using the local outlier factor algorithm \cite{breunig2000lof}). %
Since the first turbine presents a normal power curve only ($l = 1$) there is a total of five tasks, 
$\sum^L_{l=1} K_l = K_1 + K_2 = 3 + 2 = 5$. %
As before, specific tasks have less data than others, %
with the number of observations per group is as follows,

\begin{center}
\begin{tabular}{| l || c | c | c | c |}
 \hline
  & $N_{1l}$ & $N_{2l}$ & $N_{3l}$ & $ \sum_{k=1}^{K_l}$ \\
 \hline
  normal $(l=1)$ &  1075 & 1869 & 5845 & 8789 \\
  curtailed $(l=2)$ & - & 637 & 1155 & 1792 \\
 \hline
\end{tabular}
\end{center}

The proportions of training data are listed below. %
The observations remain ordered to test generalisation to measurements from later operational life. %

\begin{center}
\begin{tabular}{| c | c | c |}
 \hline
 $k=1$ & $k=2$ & $k=3$ \\
 \hline
 90\% & 66\% & 66\% \\
 \hline
\end{tabular}
\end{center}

The splits are intentionally inconsistent, to allow a
combined inference to leverage information from the data-rich tasks (with historical data) to support sparse tasks (systems recently in operation). %
In particular, referring to \Cref{f:PC-data}: the normal data from the first turbine ($k=1, l=1$: dark blue) should support the sparse normal tasks ($k \in \{2, 3\}$: dark orange and green); while the data-rich curtailment from the third turbine (light green) should support the curtailed relationship of the second turbine (light orange). %

\subsection{Task regression formulation}
A standard power curve model assumes segmented linear regression~\cite{lydia2014comprehensive}. A similar formulation is adopted here, %
\begin{align}
P(x_i) &= \begin{cases}
    0 & x_i < p\\
    m_1 (x_i - p) & p < x_i < q\\
    m_2 (x_i - q)  + m_1( q - p) & q < x_i < r\\
    P_m & x_i > r
    \end{cases} \nonumber\\[1em]
    m_2 &\triangleq \frac{P_m - m_1( q - p)}{(r - q)} \label{e:s-pw}
\end{align}
Although simple, (\ref{e:s-pw}) presents interpretable parameters -- visualised in \ref{a:vis-pwLin}. %
$p$~is the cut-in speed and $r$ is the rated speed (for normal operation);
the change-point $q$ corresponds to the initiation of the limit to maximum power $P_m$ (where $p < q < r$).
The gradients $m_1$ and $m_2$ approximate the near-linear response between $p$-$q$ and $q$-$r$ respectively. %
The second change point and gradient $\{q, m_2\}$ enable \textit{soft} curtailments, rather than a hard-limit at maximum power $P_m$. %

\subsection{Mixed-effects and prior formulation}
From knowledge of turbine operation, the expected power before cut-in should be zero for all turbines (i.e.\ a fixed effect). %
The cut-in speed $p$ is also tied as a fixed effect and learnt at the population level since all turbines have the same design. %
Similarly, the max power $P_m$ is tied between operational labels $l \in \{1, 2\}$ such that one parameter is learnt for the normal tasks ($l=1$) and one for the curtailed tasks ($l=2$). %
Conversely, the change-points $\{q, r\}$ and gradients $\{m_1, m_2\}$ are assumed to be correlated between all tasks, i.e.\ correlated via shared parent nodes. %
In turn, one would expect the curtailed relationships ($l=2$) to be more correlated (and share more information) than the normal relationships ($l=1$) and vice versa. %

The (expected) tasks are summarised as segmented mixed effects,

\begin{footnotesize}
\begin{align}
&\bigg\{\big\{ \hat{y}^{(kl)}_i = \ldots \nonumber\\ 
&\begin{cases}
{\color{teal} 0} & x_i < {\color{teal} p}\\
{\color{violet} m^{(kl)}_1} (x_i - {\color{teal} p}) & {\color{teal} p} < x_i < {\color{violet} q^{(kl)}}\\
{\color{violet} m_2^{(kl)}} (x_i - {\color{violet}{q^{(kl)}}}) + {\color{violet} m_1^{(kl)}}( {\color{violet}q^{(kl)}} - {\color{teal} p })& {\color{violet}{q^{(kl)}}} < x_i < {\color{violet}{r^{(kl)}}} \\
{\color{teal} P}^{{\color{teal}(l)}}_{\color{teal} m}  & {\color{violet}{q^{(kl)}}} < x_i < {\color{violet}{r^{(kl)}}} \\
\end{cases}  \nonumber\\
& \qquad \qquad \ldots \big\}_{k=1}^{K_l}\bigg\}_{l=1}^L
\end{align}
\end{footnotesize}

\begin{align}
&{\color{violet} m_2^{(kl)}} \triangleq \frac{{\color{teal} P}^{{\color{teal}(l)}}_{\color{teal} m} - {\color{violet} m_1^{(kl)}}( {\color{violet}q^{(kl)}} - {\color{teal} p })}{({\color{violet}{r^{(kl)}}} - {\color{violet}{q^{(kl)}}})} 
\label{e:pw-mem}
\end{align}

where the fixed effects are {\color{teal} green} and the random effects are {\color{violet}{purple}}. %
Each segment of the regression could be presented in a similar formulation to (\ref{e:mem-model-eqns2}) such that each component is a standard varying intercepts/slope model \cite{gelman2013bayesian}. %
Matrix notation is avoided, however, to present the model (and priors) around parameters $\{P_m, m_1, m_2, p, q, r\}$. %
The likelihood of the response can be specified using~(\ref{e:pw-mem}),
\begin{align}
y_{ikl}| x_{ikl}, \boldsymbol{\theta}_{kl} \sim \textrm{N}\left(\hat{y}^{(kl)}_i,\; \sigma^2\right) \label{eq:lik-3}
\end{align}

where $\boldsymbol{\theta}_{kl} = \{P^{(l)}_m, m^{(kl)}_1, p, q^{(kl)}, r^{(kl)}\}$ is the parameter set indexed to turbine $k$ and curtailment $l$. %

Given their interpretability, weakly informative priors are postulated for each parameter. %
For the change points,
{\small
\begin{align}
p \sim &\textrm{N}(\mu_{p}, \sigma^2_{cp}), \quad q^{(kl)} \sim \textrm{N}(\mu_{q}, \sigma^2_{cp}), \quad r^{(kl)} \sim \textrm{N}(\mu_{r}, \sigma^2_{cp}) \nonumber\\[1em]
&\mu_p \sim \textrm{N}(.2, .5), \quad \mu_q \sim \textrm{N}(.4, .5), \quad \mu_r \sim \textrm{N}(.6, .5)  \nonumber\\
& \hspace{8em} \sigma_{cp} \sim \textrm{IG}(1, 1) \label{e:wt-prior1}
\end{align}}These priors reflect that change points are expected to occur at regular intervals across the input domain with relatively high variance (relative to a normalised scale). %
The priors for gradient and maximum power are,
\begin{align}
m_1^{(kl)} \sim \textrm{N}( &{\mu}_{m_1}, {\sigma}^2_{m_1}) \nonumber\\[1em]
 {\mu}_{m_1} \sim \textrm{N}\left(2.5, .5\right), \quad & {\sigma}_{m_1} {\sim} \textrm{IG}\left(1, 1\right)
\end{align}
\begin{align}
 P^{(1)}_m \sim \textrm{N}(1, .1), \qquad P^{(2)}_m \sim \textrm{N}(.8, .1)  \label{e:wt-prior2}
\end{align}

These distributions postulate the expected gradient $m_2$ in a normalised space; unit max power $P^{(1)}_m$ for normal operation; and a typical 80\% curtailment \cite{bull2021bayesian} for the limited output $P^{(2)}_m$. %
No prior is required for $m_2$ since it is specified by $\{P_m, m_1, p,q,r\}$ in~(\ref{e:pw-mem}). %
As with the truck-fleet example, the $\textrm{IG}(1,1)$ distributions weakly encourage inter-task correlations, such that the prior intentionally overestimates the deviation between task parameters. %
Similarly, the posterior can be specified using (\ref{e:post-1}), where $p([\mathbf{y}_k] \mid \mathbf{\Theta})$ is indexed by (\ref{eq:lik-3}) and the joint prior $p(\mathbf{\Theta})$ is defined using (\ref{e:wt-prior1}) to (\ref{e:wt-prior2}). %
As before, this is intractable and inferred with MCMC. %

\subsection{Results}
\begin{figure*}[ht]
    \centering\includegraphics[width=\linewidth]{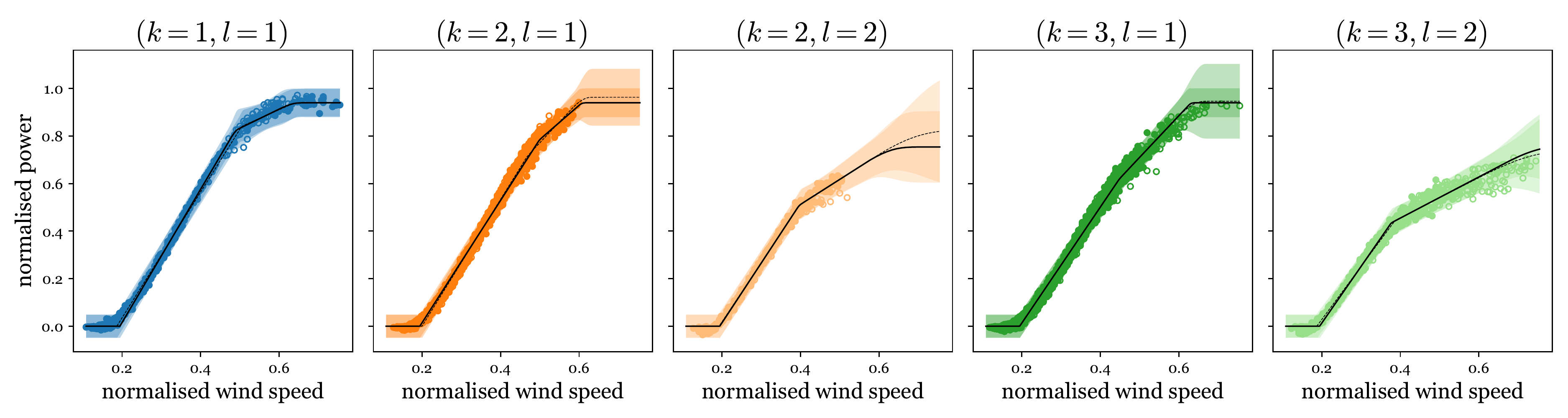}
    \caption{Posterior predictive distribution, the mean and three-sigma deviation for: (light shading, dashed line) $K$ independent power-curve models $p(\mathbf{y}^*_{kl} | \mathbf{x}_{kl}^*, \mathbf{x}_{kl}, \mathbf{y}_{kl})$. (dark shading, solid line) multitask learning with mixed effects $p(\mathbf{y}^*_{kl} | \mathbf{x}_{kl}^*, \{\{\mathbf{x}_{kl}, \mathbf{y}_{kl}\}_{k=1}^{K_l}\}_{l=1}^L)$.} %
    \label{f:PC-funcs}
\end{figure*}

\Cref{f:PC-funcs} shows posterior predictive distribution from fleet-level inference -- compared to independent STL models, plotted with light shading. %
Intuitively, variance reduction is most obvious for sparse or poorly described domains (orange and dark green). %
There is an overall increase in the predictive likelihood when fleet modelling, compared to single-task learning, from 8229 to 8258. %
\Cref{t:wt-L} quantifies changes in task-wise predictions compared to the benchmarks: there is a likelihood increase in all domains other than ($k=2, l=1$) and ($k=3, l=2$). %
It is believed that reductions occur since the model is constrained such that, to maximise the overall likelihood, the performance in data-rich domains is reduced in a trade-off. %
In other words, the prior belief is best suited to data-rich tasks -- when the prior becomes more informed by data, it becomes less suitable in data-rich domains; %
instead, the prior represents the population. %
(Consider that the overall likelihood $\mathcal{L}$ increases, despite task-wise fluctuations.) %
To combat this, uninformative priors should be considered \cite{gelman2013bayesian}; these are discussed in \Cref{s:conc}. %

Correlation alignment (CRL) performs less competitively in the wind turbine example since the measurement distributions shift significantly between each task, training, and testing (testing data correspond to following weeks). %
In particular, when the source data represent a more complete power curve, the alignment with sparse domains becomes partial, and CRL can produce unreasonable embeddings. %

\begin{table}
\begin{center}
\caption{Predictive log-likelihood $\log p(\mathbf{y}^*_{kl} | \mathbf{x}^*_{kl})$. Here $l$ corresponds to the operating condition (normal $l=1$, or curtailed $l=2$) while $k$ is turbine identifier.}
\resizebox{\linewidth}{!}{%
\begin{tabular}{ | c || c | c | c | c | c | c | c | c }
 \hline
  method & $\makecell{k=1,\\ \;\;l=1}$ & $\makecell{k=2,\\ \;\;l=1}$ & $\makecell{k=3,\\ \;\;l=1}$ & $\makecell{k=2,\\ \;\;l=2}$ &  $\makecell{k=3,\\ \;\;l=2}$ &$\mathcal{L}$ \\
\hline
  CP  & -168 & 1555 & 4681 & 594 & 452 & 7114  \\

  CRL & 52 & 1451 & 3359 & 282 & -6 & 5138 \\

  STL & 202 & \textbf{1619} & 5147 & 538 & \textbf{722} & 8229 \\

  \textit{MTL} & \textbf{218} & 1599 & \textbf{5206} & \textbf{549} & 686 & \textbf{8258}\\
  \hline
\end{tabular}}\label{t:wt-L}
\end{center}
\end{table}

\begin{figure*}[ht]
    \centering\includegraphics[width=\linewidth]{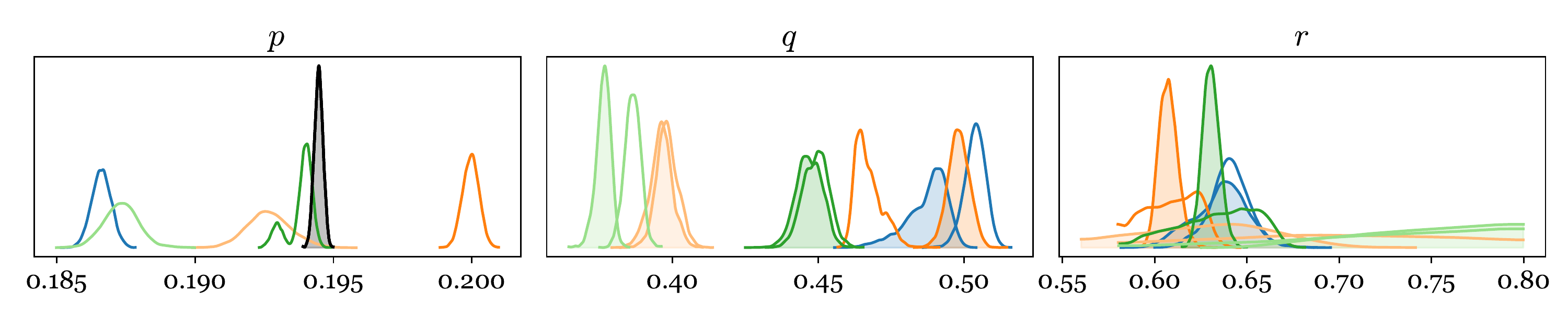}
    \includegraphics[width=\linewidth]{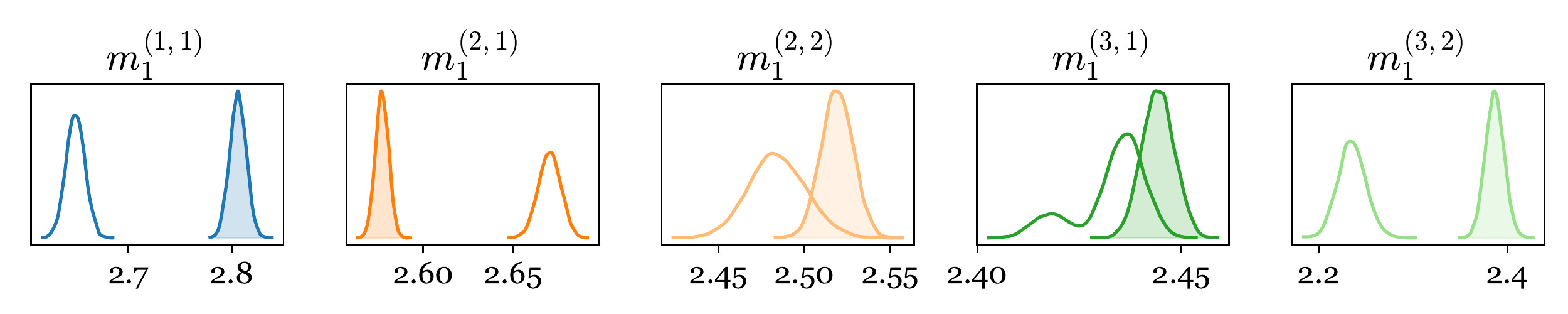}
    \caption{Changes in the posterior distribution: the cut-in speed $p$, initiation of curtailment $q$, rated speed $r$, and linear slope $m_1$. Independent models (hollow) compared to population-level modelling (shaded). When the parameter is tied (or fixed) the distributions are black.} %
    \label{f:VR-PCs}
\end{figure*}

\Cref{f:VR-PCs} shows the posterior distribution of the parameters inferred at the independent and fleet level. %
The cut-in speed $q$ moves towards an average of the independent models, with reduced variance; this should be expected since $q$ becomes tied as a population estimate. %
The change points $q$ cluster intuitively, such that the normal and curtailed tasks form two groups (dark and light shades). %
The estimated $r$ parameters are significantly improved through partial pooling -- in particular, the green and orange domains shift much further from the weakly informative prior. %
There is a notable reduction in the variance across all tasks for the slope estimate~$m_2$. %
The average reduction in standard deviation across these parameters is 25\%. 

\begin{figure}[ht]
    \centering \includegraphics[width=.9\linewidth]{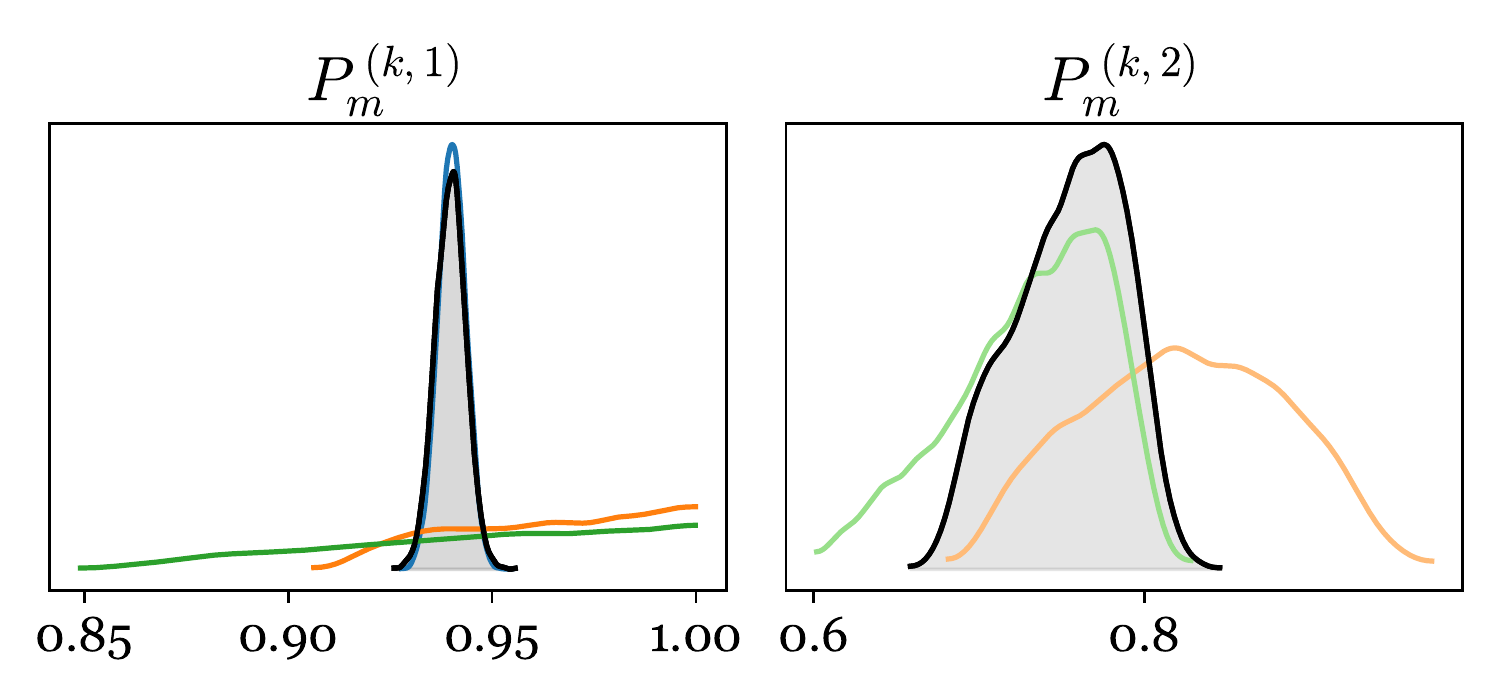}
    \caption{Changes in the posterior distribution of the:  Independent models (hollow) compared to population-level modelling (shaded).} %
    \label{f:VR-Pms}
\end{figure}

\Cref{f:VR-Pms} presents insights relating to maximum power estimates $P_m$. %
The tied parameter for the normal maximum $P^{(k,1)}_m$ moves toward the data-rich estimate (blue) while the curtailed maximum $P^{(k,2)}_m$ moves toward an average of the relevant tasks (where $l=2$). %
In both operating conditions, parameter tying enables the move from vague posteriors to distributions with clear expected values. %
The average reduction in standard deviation for the normal maximum is 82\%, alongside 37\% for the curtailed maximum. %

\begin{figure}[ht]
    \includegraphics[width=\linewidth]{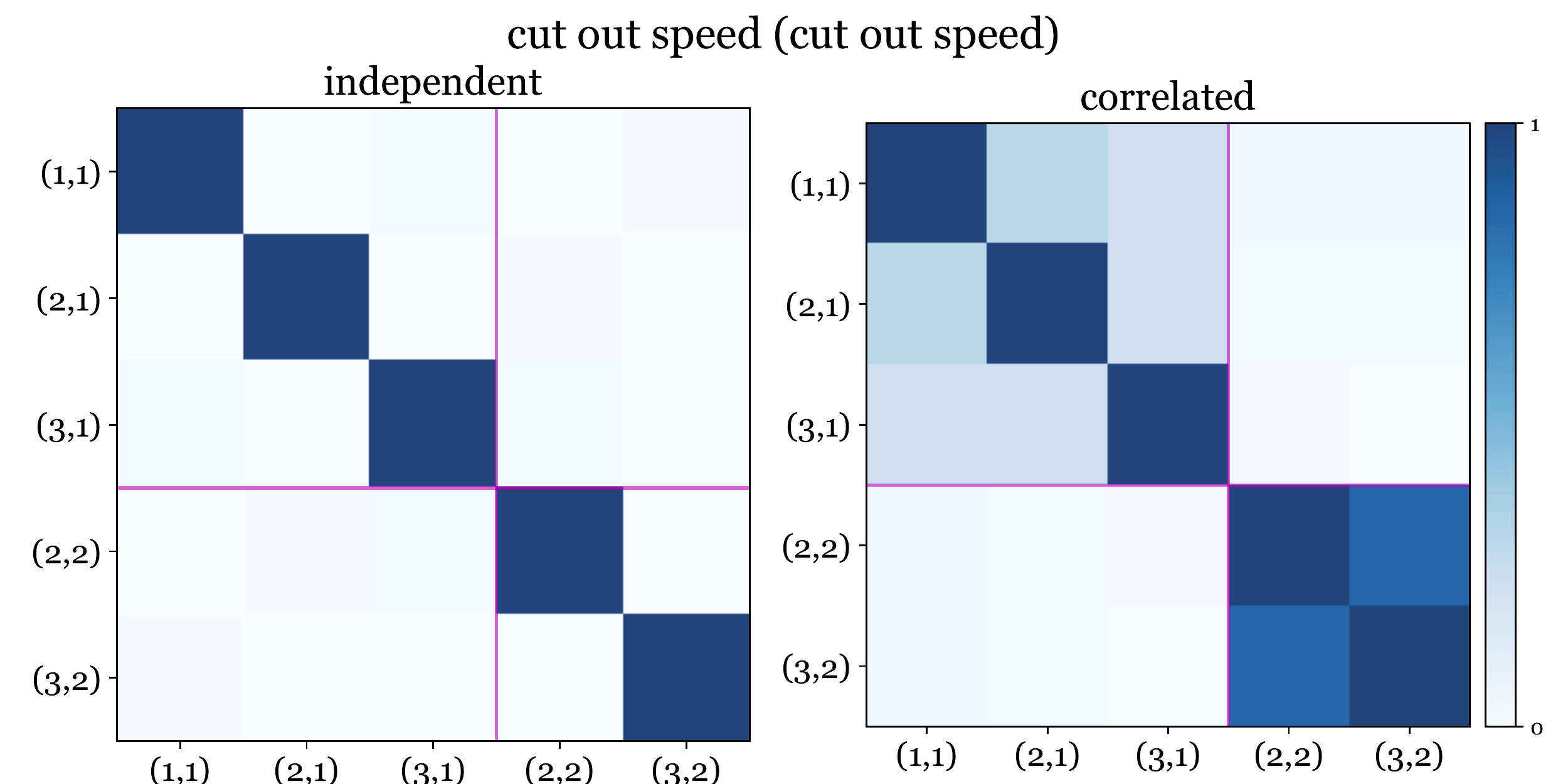}
    \caption{Pearson correlation coefficient of the conditional posterior distribution for the rated wind speed $q$, %
    tick labels correspond to ($k,l$). Purple lines separate the normal ($l=1$) from the curtailed task-parameters ($l=2$).} %
    \label{f:Rq}
\end{figure}

Finally, \Cref{f:Rq} plots the Pearson correlation coefficient of the pair-wise conditionals of $q$ between tasks. ($q$ is presented since it is the most structured/insightful.) %
It is clear that, by moving to a hierarchical model, the correlation between related tasks is appropriately captured, with two distinct blocks associated with the normal and curtailed groups. %

\subsection{Practical implications: Decision analysis}\label{s:decisions}

In practice, probabilistic predictions from the power model can be used to support decisions at any level of the hierarchy, including the population level. %
For example, population-level decisions are useful if the operator does not wish to commit to interacting with a specific turbine.

\begin{figure}
    \centering
    \includegraphics[width=.8\linewidth]{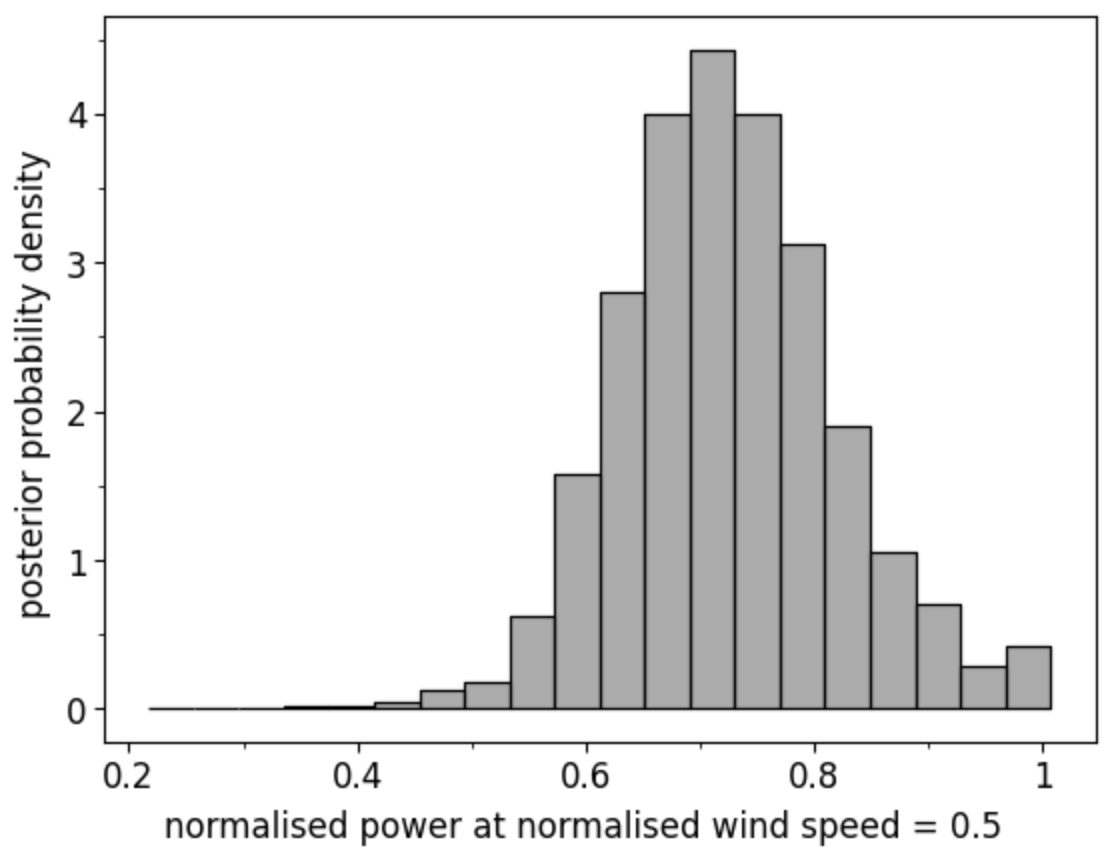}
    \caption{Samples from the posterior predictive distribution of normalised power, at an arbitrary input of normalised wind speed = $0.5$.}
    \label{fig:post_pred_example}
\end{figure}

Consider a decision problem, whereby an operator must commit to delivering a minimum power in some upcoming time window. %
This involves decision making under uncertainty, and the formal (statistical) procedure to identify the expected optimal action requires a probabilistic quantification of wind speed and power output. %
The latter can be achieved by sampling from the posterior predictive distribution at the population level, i.e.\ $p(\mathbf{y}^* \mid \mathbf{x}^*,\boldsymbol{\theta}_{l})$, where $\boldsymbol{\theta}_{l} = \{P^{(l)}_m, m^{(l)}_1, p, q^{(l)}, r^{(l)}\}$ is sampled directly from the generating distributions. 
Figure \ref{fig:post_pred_example} is an example of such a prediction for a given wind speed. 

Predictions at this level of the hierarchy are useful since they assume the operator cannot commit to a specific turbine (at this stage). %
Such predictions would not be available from domain-specific (independent) models; %
conversely, complete pooling (or domain adapted) predictions would not formally consider the additional variability associated with non-specific turbine identity. %



\begin{figure}[ht]
    \centering
    \includegraphics[width=.8\linewidth]{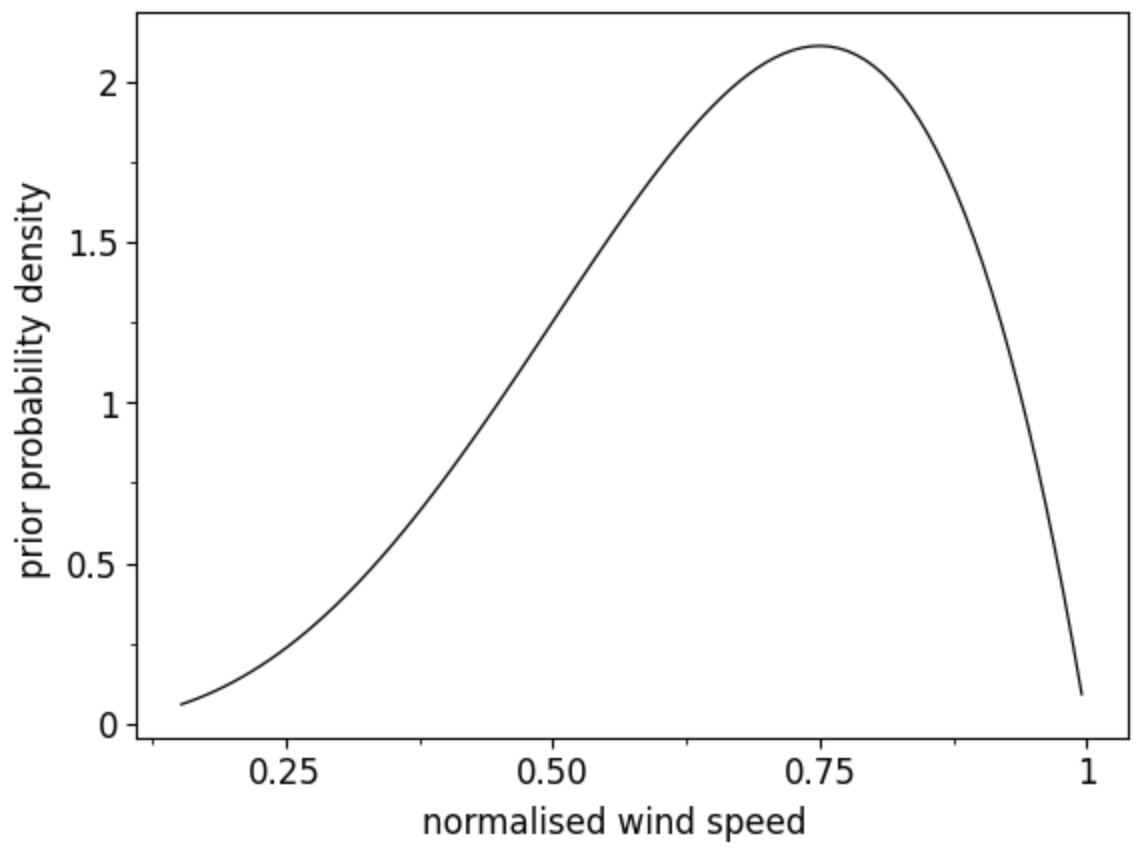}
    \caption{The prior distribution of normalised wind speed.}
    \label{fig:x_pr}
\end{figure}

In this example, the operator has three options, each associated with a payout (positive utility) upon successfully delivery of power and a penalty fine (negative utility) if the turbine generates insufficient power -- these values are presented in \Cref{t:voi}. A prior probabilistic model of (normalised) wind speed $\mathbf{x}_{pr}$ is shown in Figure \ref{fig:x_pr}, as described by,
\begin{equation}
    \mathbf{x}_{pr} \sim \textrm{Beta}(4, 2) \label{eq:x_pr}
\end{equation}
(In practice, this information would likely come from a forecasting model.) %

\begin{table}[h]
    \centering
    \caption{Financial outcomes of decision analysis.}
    \begin{tabular}{| c || c | c |}
    \hline
     Power Level & Payout & Penalty-fine\\
     \hline
        $L_{0}$: 0.0 & 0.0  & -0.0 \\
        $L_{1}$: 0.5 & 0.3 & -0.3 \\
        $L_{2}$: 0.75 & 0.75  & -1.0 \\
      \hline
    \end{tabular}
    \label{t:voi}
\end{table}

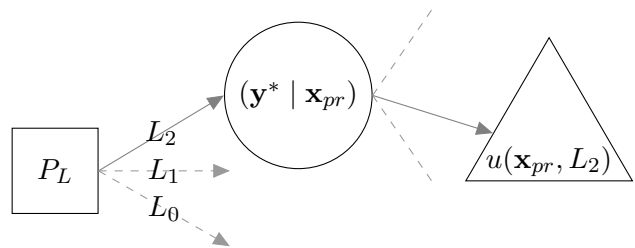
\begin{figure}
    \centering
        \begin{tikzpicture}[square/.style={regular polygon,regular polygon sides=4}]
        \node at (0,0) [square,draw, inner sep=1ex] (Pl) {$P_L$};
        \node at (3.2,1) [circle,draw] (pred) {$(\mathbf{y}^* \mid \mathbf{x}_{pr})$};
        \node at (6.5, .5) [regular polygon, regular polygon sides=3, draw, inner sep=2.7ex, label={[label distance=-12ex]:\normalsize{$u(\mathbf{x}_{pr}, {{L_2}})$}}] (u) {};

        \draw[->,draw=black!50] (Pl.east) -- node {$L_2$} (pred.west);
        \draw[dashed, ->,draw=black!40] (Pl.east) -- node {$L_1$} (2.3, 0);
        \draw[dashed, ->,draw=black!40] (Pl.east) -- node {$L_0$} (2.3, -1);
        
        \draw[dashed, draw=black!40] (pred.east) -- (5, 2.2);
        \draw[->,draw=black!50] (pred.east) -- (u.west);
        \draw[dashed, draw=black!40] (pred.east) -- (5, -0.2);
        
    \end{tikzpicture}
    \caption{Decision-event tree representation of power-level decision analysis.}
    \label{fig:prior_decision_tree}
\end{figure}

Figure \ref{fig:prior_decision_tree} shows the decision-event tree representation of the problem. %
Here, the square (decision) node $P_{L}$ is associated with the available power commitments in Table \ref{t:voi}, such that $ P_L= \{L_0, L_1, L_2\}$. The circular (probabilistic) node $(\mathbf{y}^* \mid \mathbf{x}_{pr})$ is the probabilistic prediction of power, given the prior model of wind speed. Finally, the triangular (utility) node shows the expected consequence of the decision. %

For each instance, the expected optimal action $P_{L}^{*} \in \{L_0, L_1, L_2\}$ and associated expected utility $E[u(\mathbf{x}_{pr}, P_{L}^{*})]$ are calculated,
\begin{align}
P^*_L &= \textrm{argmax}_{P_L} \textrm{E}[u(\mathbf{x}_{pr}, P_L)] \\
\textrm{E}[u(\mathbf{x}_{pr}, {P^*_L})] &= \textrm{E}[\textrm{pay}_{P^*_L}] - \textrm{E}[\textrm{penalty}_{P^*_L}]
\end{align}
where,
\begin{align}
\textrm{E}[\textrm{pay}_{P_L}] &=P(\mathbf{y}_* \geq {P_L}) \times \textrm{payout}_{P_L} \\
\textrm{E}[\textrm{penalty}_{P_L}] &= P(\mathbf{y}_* < {P_L}) \times \textrm{penaltyfine}_{P_L}
\end{align}
This information can then be used to rank decision alternatives \cite{schlaifer1961applied}. %
For example, in the prior decision tree (\Cref{fig:prior_decision_tree}) the path associated with the highest power level $L_{2}$ is optimal (i.e.\ $P_L^* = L_2$) -- this was found to have the highest expected utility of $0.33$, compared with $0.0$ for $L_{0}$, and $0.246$ for $L_{1}$.


\begin{figure*}[ht]
    \centering
        \begin{tikzpicture}[square/.style={regular polygon,regular polygon sides=4}, scale=0.85, every node/.style={scale=0.85}]
        
        \node at (-5.5, -1) [square,draw, inner sep=1ex] (M) {$M$};
        
        \node at (-3.2,1) [circle,draw, inner sep=2ex] (xm) {$\mathbf{x}_{m}$};
        \node at (0,0) [square,draw, inner sep=1ex] (Pl) {$P_L$};
        \node at (3.2,1) [circle,draw] (pred) {$(\mathbf{y}^* \mid \mathbf{x}_{m})$};
        \node at (6.5, .5) [regular polygon, regular polygon sides=3, draw, inner sep=2.7ex, label={[label distance=-12ex]:\normalsize{$u(\mathbf{x}_{m}, {{L_2}})$}}] (u) {};

        \draw[->,draw=black!50] (Pl.east) -- node {$L_2$} (pred.west);
        \draw[dashed, ->,draw=black!40] (Pl.east) -- node {$L_1$} (2.3, 0);
        \draw[dashed, ->,draw=black!40] (Pl.east) -- node {$L_0$} (2.3, -1);
        
        \draw[dashed, draw=black!40] (pred.east) -- (5, 2.2);
        \draw[->,draw=black!50] (pred.east) -- (u.west);
        \draw[dashed, draw=black!40] (pred.east) -- (5, -0.2);
        
        \draw[dashed, draw=black!40] (xm.east) -- (-1.4, 2.2);
        \draw[->,draw=black!50] (xm.east) -- (Pl.west);
        \draw[dashed, draw=black!40] (xm.east) -- (-1.4, -0.2);
        \draw[->, draw=black!50] (M.east) -- node {${m}$} (xm);
        
        \node at (-3.2,1 - 4) [circle,draw, inner sep=2ex] (xpr) {$\mathbf{x}_{pr}$};
        \node at (0,0 - 3) [square,draw, inner sep=1ex] (Pl2) {$P_L$};
        \node at (3.2,1 - 3) [circle,draw] (pred2) {$(\mathbf{y}^* \mid \mathbf{x}_{pr})$};
        \node at (6.5, .5 - 3) [regular polygon, regular polygon sides=3, draw, inner sep=2.7ex, label={[label distance=-12ex]:\normalsize{$u(\mathbf{x}_{pr}, {{L_2}})$}}] (u2) {};

        \draw[->,draw=black!50] (Pl2.east) -- node {$L_2$} (pred2.west);
        \draw[dashed, ->,draw=black!40] (Pl2.east) -- node {$L_1$} (2.3, 0 - 3);
        \draw[dashed, ->,draw=black!40] (Pl2.east) -- node {$L_0$} (2.3, -1 - 3);
        
        \draw[dashed, draw=black!40] (pred2.east) -- (5, 2.2 - 3);
        \draw[->,draw=black!50] (pred2.east) -- (u2.west);
        \draw[dashed, draw=black!40] (pred2.east) -- (5, -0.2 - 3);
        
        \draw[->,draw=black!50] (xpr.east) -- (Pl2.west);
        \draw[->, dashed, draw=black!40] (M.east) -- node {$\bar{m}$} (xpr);

    \end{tikzpicture}
    \caption{Decision-event tree representation of the value of information analysis.} %
    \label{fig:voi_decision_tree}
\end{figure*}
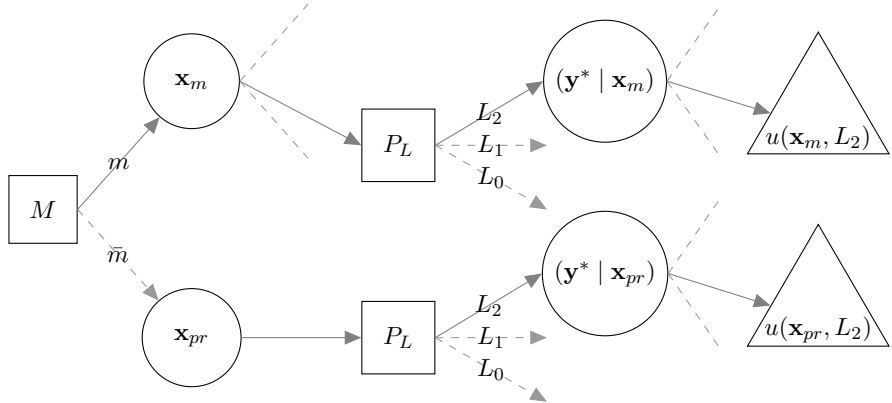

A further application quantifies the expected value of data collection activities. %
Figure \ref{fig:voi_decision_tree} extends the problem in Figure \ref{fig:prior_decision_tree} to include another decision $M$: whether to measure wind speed ($m$) or not ($\bar{m}$). %
In the case where measurements are taken, predictions can be made using the new data $\textbf{x}_{m}$.  %
A so-called \textit{preposterior} decision analysis \cite{berger2013statistical,jordaan2005decisions} can be completed, by sampling from the prior model to generate hypothetical measurements. 

When assuming \emph{perfect} data, whereby each measurement removes all uncertainty from wind speed (\ref{eq:x_pr}), the expected (preposterior) utility is $0.566$. %
The difference in expected utility -- with ($m$) or without ($\bar{m}$) wind measurement -- is the expected value of the data $\textbf{x}_{m}$ in the context of solving the decision problem. This expected Value of Perfect Information (VoPI) can be estimated using Monte Carlo sampling, %
\begin{equation}
    VoPI = \frac{1}{N} \sum_{i = 1}^{N} \Big( E[u(\textbf{x}_m, P_{L}^{*})] \Big) - E[u(\mathbf{x}_{pr},  P_{L}^{*})]
    \label{eq:VoPI}
\end{equation}  

\begin{figure}
    \centering
    \includegraphics[width=.9\linewidth]{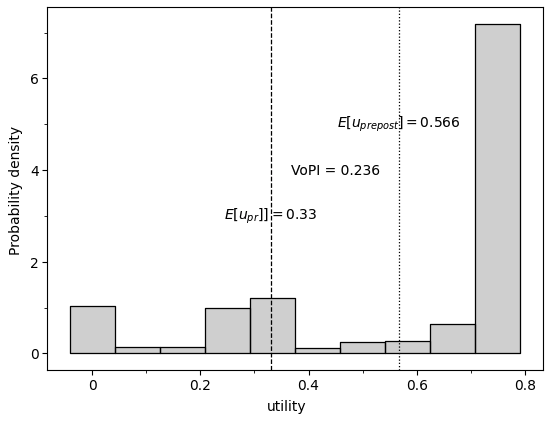}
    \caption{Expected utilities associated with hypothesised wind speed measurements. The expected VoPI is shown as the difference between expected utilities with ($E[u_{prepost}]$) and without ($E[u_{prior}]$) the wind speed data.}
    \label{fig:voi_results}
\end{figure}

Here, the VoPI is $0.236$. 
The results are presented in Figure \ref{fig:voi_results}, which shows a histogram of expected utilities associated with each of the hypothesised, perfect measurements (samples from the prior model). The mean of these values $E[u_{prepost}]$ is labelled next to the dotted line. The expected utility without the data $E[u_{pr}]$ is labelled as a dashed line, and the difference (\ref{eq:VoPI}) is the expected value of the data.


To summarise, hierarchical Bayesian modelling has provided a full quantification of uncertainty, reflective of different asset subgroups and classes. %
In turn, the model enables a formal downstream analysis of variable interactions and integration with a utility-based decision process (demonstrated here). %
The implications are significant since various concepts can be quantified; for example, the expected \textit{optimal action} or the \textit{value of data collection} activities. %


\section{Concluding Remarks}\label{s:conc}

Hierarchical Bayesian modelling with mixed effects is demonstrated as an effective method of sharing information between models of fleets of assets in engineering. 
Parameter estimation and predictive capabilities are improved (for the combined fleet) in two case studies, utilising the same flexible multi-task learning framework. %
Important considerations are discussed when formulating each population model: prior elicitation, mixed-effects formulation, and negative transfer -- these concepts are critical to the success of population-level inference. %

The proposed hierarchical methodology is desirable since it enables downstream analyses of the fleet model. %
The method is used to determine which asset models are correlated for which interpretable parameter, at various groupings (e.g.\ operating condition, system-specific, population-wide). %
The multivariate (and multilevel) uncertainty quantification enabled by the model is then propagated through a demonstrative decision analysis for the second case study, to consistently and coherently identify expected optimal actions. %
The expected value of data collection is also quantified, in the context of the decision problem and the underlying model. %

The first application concerns the survival analysis of turbocharger and alternator components in an operational fleet of trucks (maintained by Scania). %
A semi-parametric hazard curve model is improved through partial pooling and parameter tying (15\% and 13\% increases in predictive log-likelihood) where selected parameters are inferred at the population-level, rather than vehicle subgroups. %
The method builds on engineering intuition since correlations in the hierarchy can be inspected to determine which groups of vehicles or components are correlated for which effects in the survival model (i.e.\ interpretable parameters). %

The second study presents power prediction for a group of wind turbines. %
The SCADA monitoring data were provided by Visualwind, measured from the same model of turbine in different locations. %
Correlated power curve models are learnt as a segmented (piece-wise) linear regression, described by interpretable parameters. %
By moving to a population-level inference, parameter estimation is improved, as well as model generalisation (for the combined population estimates). %
In particular, the estimation of maximum power is significantly improved for turbines with fewer data and recently in operation (up to 82\% reduction in the standard deviation of maximum output prediction). %

The success of these models depends on the reliability of the domain knowledge encoded in the prior distributions. %
In this case, priors were postulated as weakly informative, since interpretable parameters and domain expertise allowed sensible prior elicitation. %
In turn, an appropriate level of knowledge transfer could be determined automatically, given the model and the data, reducing the risk of negative transfer. %
When such elicitation is infeasible, future work should consider the use of uninformative priors \cite{gelman2006prior}, especially for the (variance) parameters that control the level of correlation between tasks. %

Future work should consider an objective method to categorise sub-fleet data in a practical setting; this might include clustering assets from specification or operations data. %
The labelling of data into distinct tasks can be non-trivial in an engineering setting and requires investigation. %
Finally, extending the multilevel model to capture parameter relationships over the fleet should prove insightful; for example, if the coefficients of the power model were regressed on spatial/temporal inputs for the wind farm, one could simulate (sample) more varied hypothetical members of the population at certain locations or timescales. %

\section*{Acknowledgements}

A.B. Duncan, D. Di Francesco and L.A.\ Bull were supported by Wave 1 of The UKRI Strategic Priorities Fund under the EPSRC Grant EP/W006022/1, particularly the \textit{Ecosystems of Digital Twins} theme within that grant and The Alan Turing Institute. %
M.\ Dhada was supported by the Next Generation Converged Digital Infrastructure project (EP/R004935/1) funded by the Engineering and Physical Sciences Research Council and BT. 
This research was supported by Scania CV (Sweden) and Visualwind (UK). %
The authors would also like to thank Dr.\ Paul Gardner and Jack Poole for their helpful conversations while writing this paper.

\bibliographystyle{unsrtnatemph}
\bibliography{ref}

\appendix

\section{B-splines}\label{a:b-splines}
Assuming uniform knot locations $x_{h+k} = x_h + \delta k$, cubic B-splines are defined as the following piece-wise cubic polynomial \cite{gelman2013bayesian},
\begin{tiny}
\begin{align}
b_h(x) =
\begin{cases}
\frac{1}{6} u^3 & x \in (x_h, x_{h+1}), \quad u = (x- x_h)/\delta \\
\frac{1}{6} ( 1 + 3u + 3u^2 - 3u^3) & x \in (x_{h+1}, x_{h+2}), \quad u = (x- x_{h+1})/\delta \\
\frac{1}{6} (4 - 6u^2 + 3u^3) & x \in (x_{h+2}, x_{h+3}), \quad u = (x- x_{h+2})/\delta \\
\frac{1}{6} (1 - 3u + 3u^2 - u^3) & x \in (x_{h+3}, x_{h+4}), \quad u = (x- x_{h+3})/\delta \\
0 & \textrm{otherwise} 
\end{cases}
\end{align}
\end{tiny}

\section{Cross validation Scania}\label{a:cv-scania}
\begin{figure}[h]
    \centering\includegraphics[width=.65\linewidth]{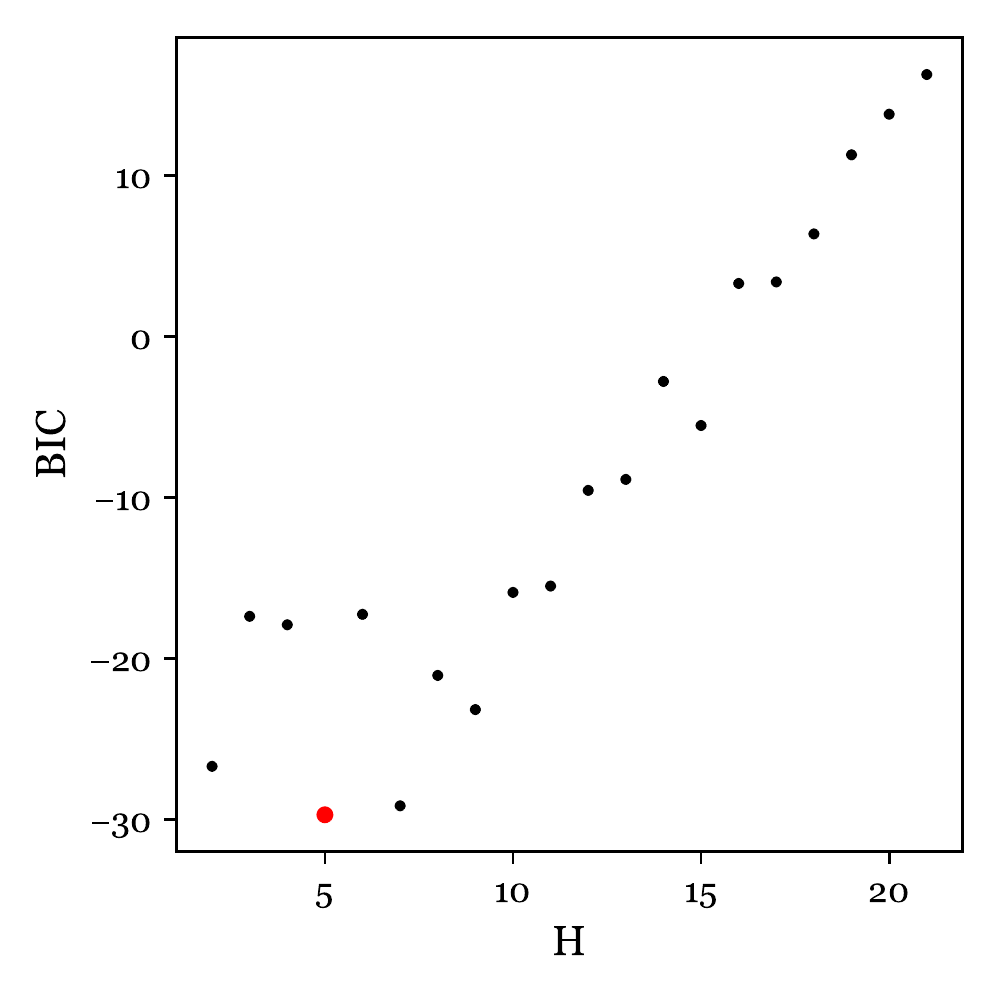}
    \caption{Validation of an appropriate number of splines using the Bayesian Information Criterion (BIC) and 20-fold cross-validation. The best model $H=5$ is highlighted with a red marker.}
    \label{f:cv-splines}
\end{figure}

\newpage
\section{Segmented (piece-wise) power curve model}\label{a:vis-pwLin}
\begin{figure}[h]
    \centering\includegraphics[width=.8\linewidth]{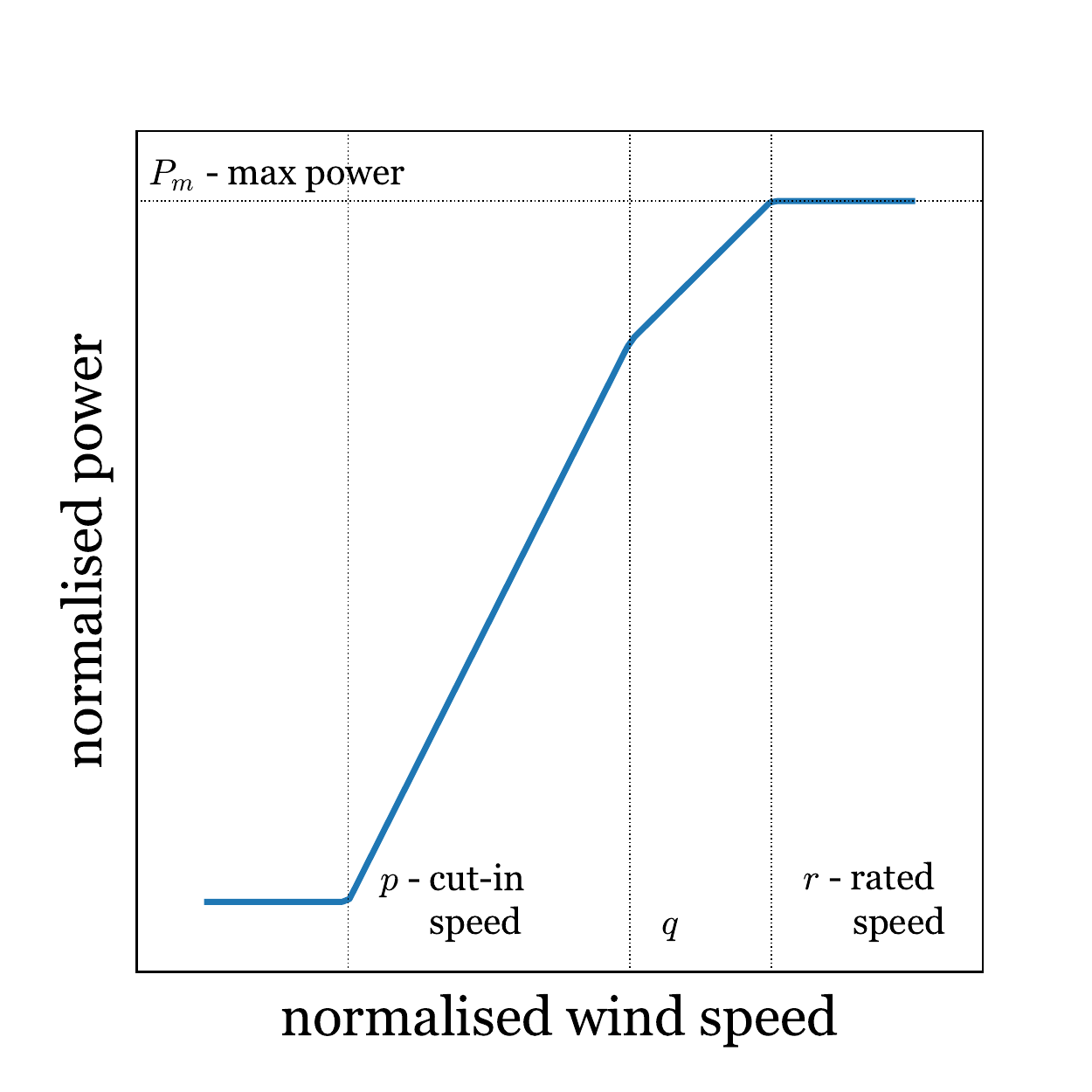}
    \caption{The segmented linear power-curve model, indicating interpretable parameters $\{p, q, r, P_m\}$.}
    \label{f:vis-pwLin}
\end{figure}

\section{Zoomed spline weights}\label{a:zoomed}
\begin{figure}[h]
    \centering\includegraphics[width=.8\linewidth]{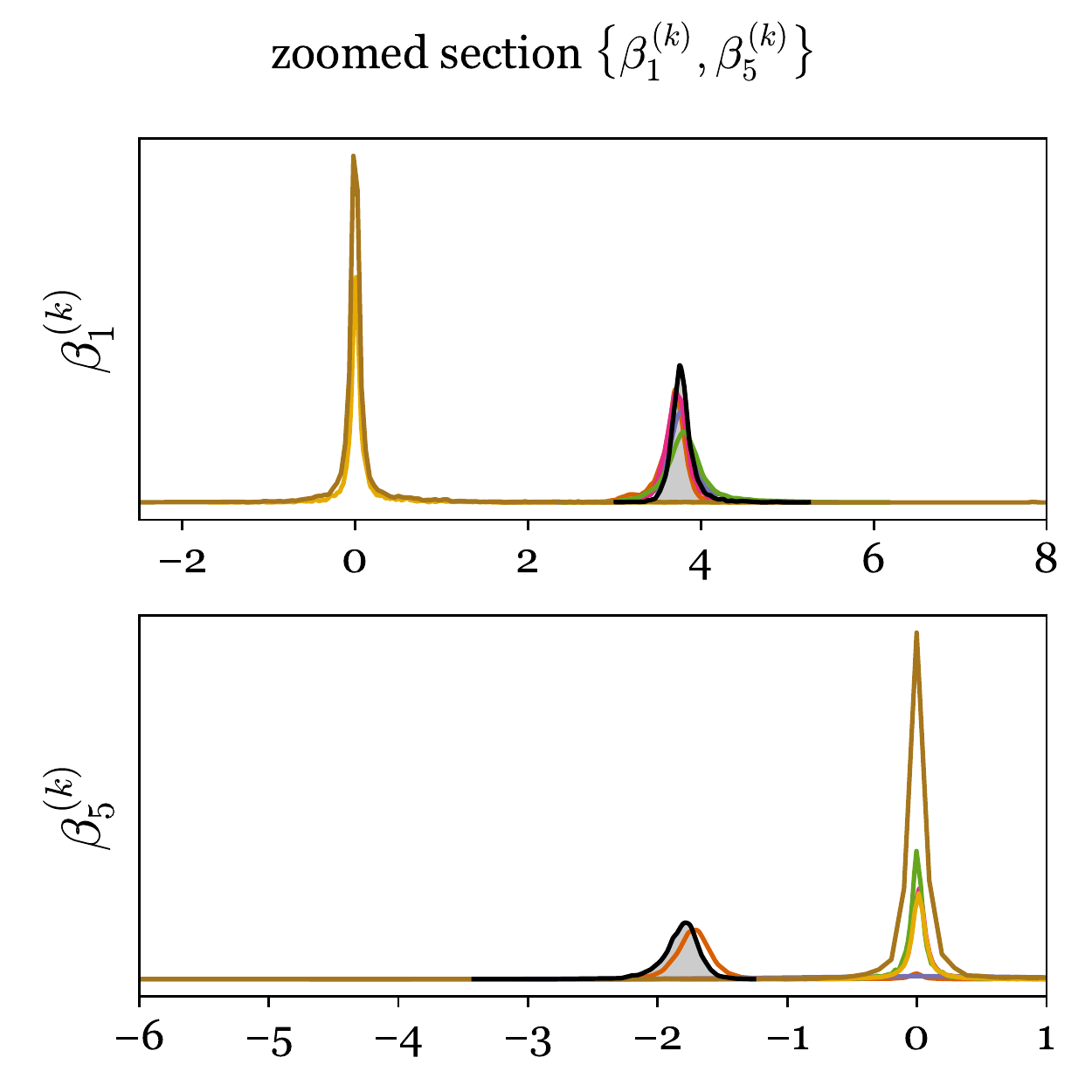}
    \caption{Zoomed sections of the posterior distribution of the spline weights $\beta_h^{(k)}$ (those that deviate from zero $h \in \{1, 5\}$).}
\end{figure}

\newpage
{\onecolumn

\section{Turbocharger model: consistent model formulation}\label{a:tc-initial}
\begin{figure*}[ht]
    \centering\includegraphics[width=.7\linewidth]{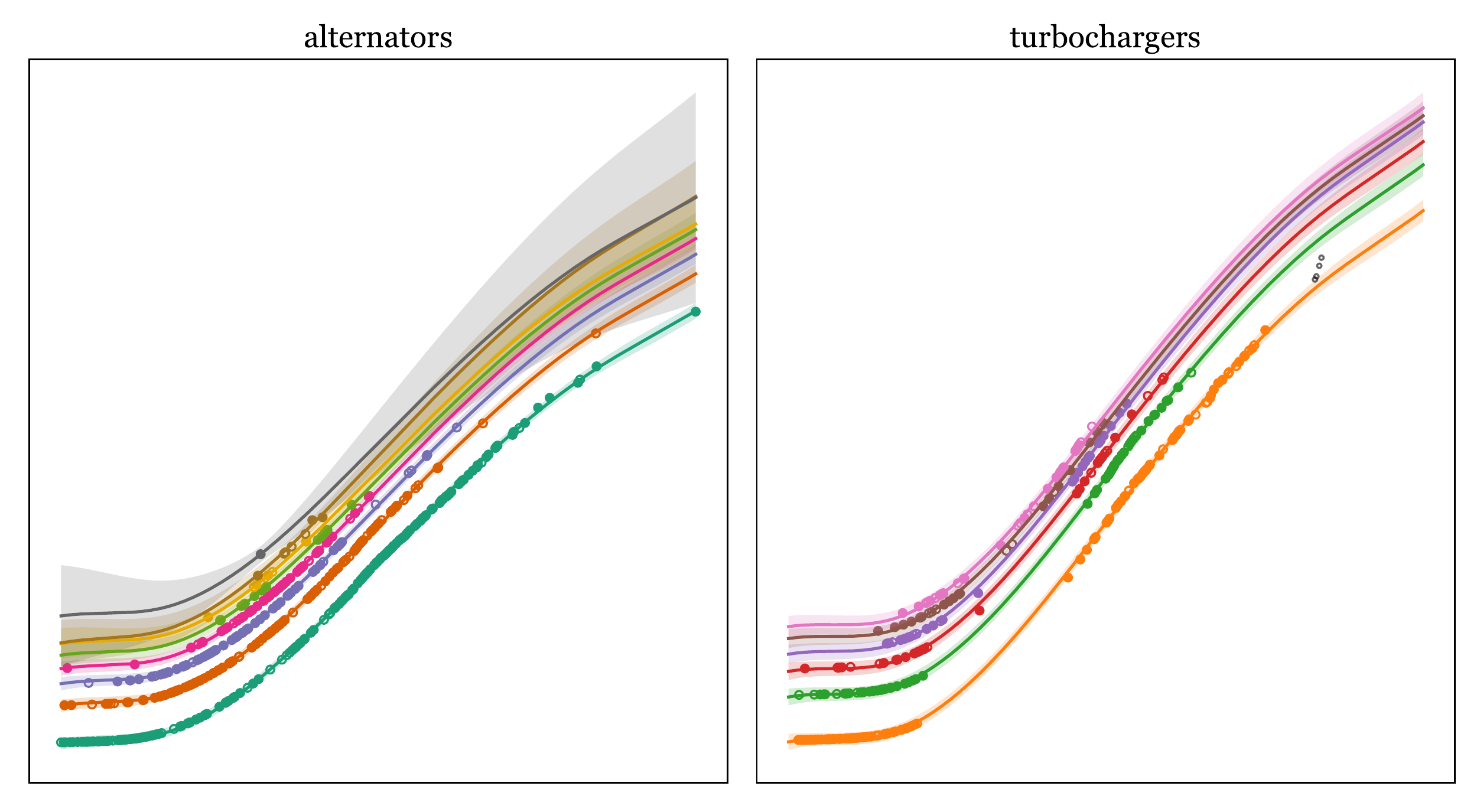}
    \caption{Posterior predictive distribution $p(\mathbf{y}^*_k | \mathbf{x}_k^*, \{\mathbf{x}_k, \mathbf{y}_k\}_{k=1}^K)$: the mean and three-sigma deviation for multitask learning with mixed effects.} %
    \label{f:all-funcs-MEM}
\end{figure*}

\section{Turbocharger model: variance reduction plots}\label{a:tc-VR}
\begin{figure*}[h]
    \centering\includegraphics[width=.6\linewidth]{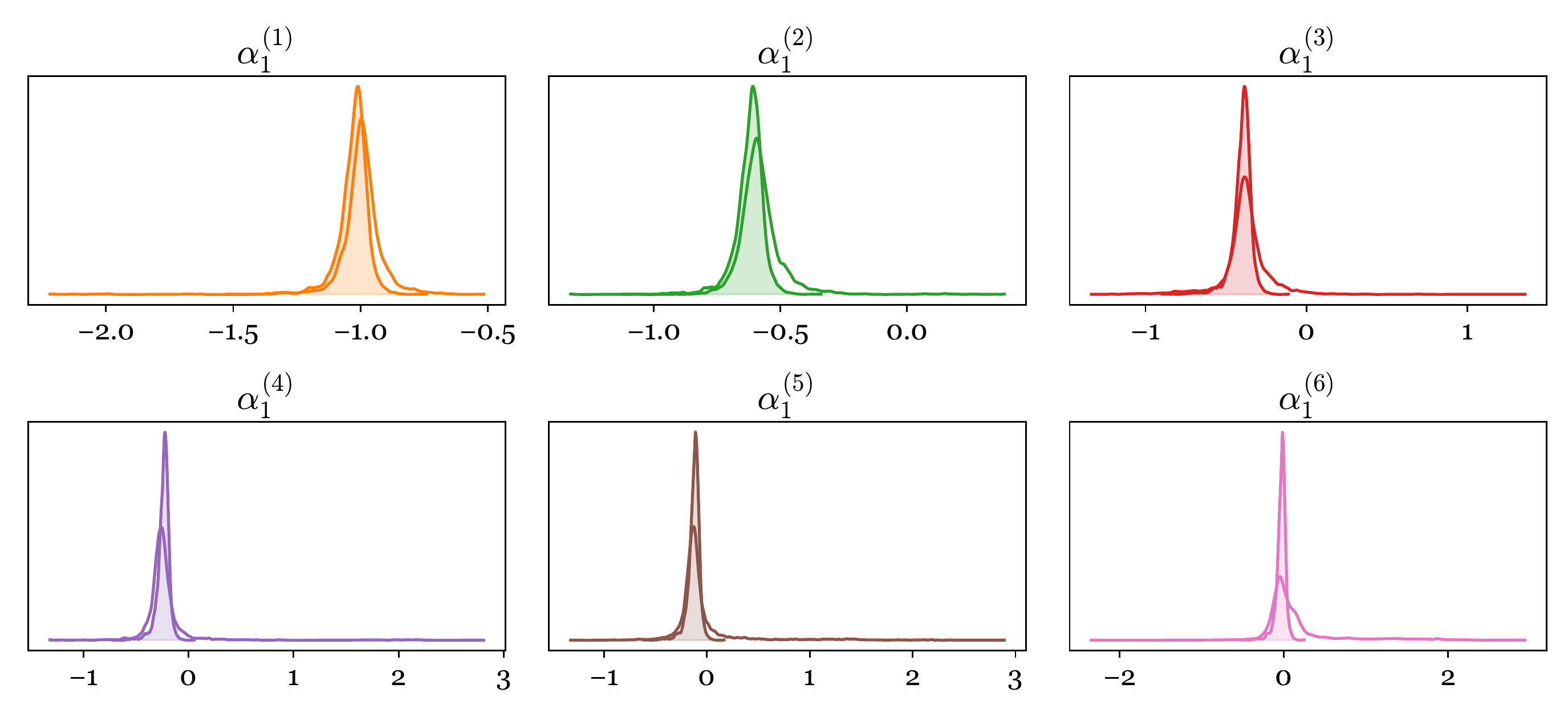}
    \centering\includegraphics[width=.6\linewidth]{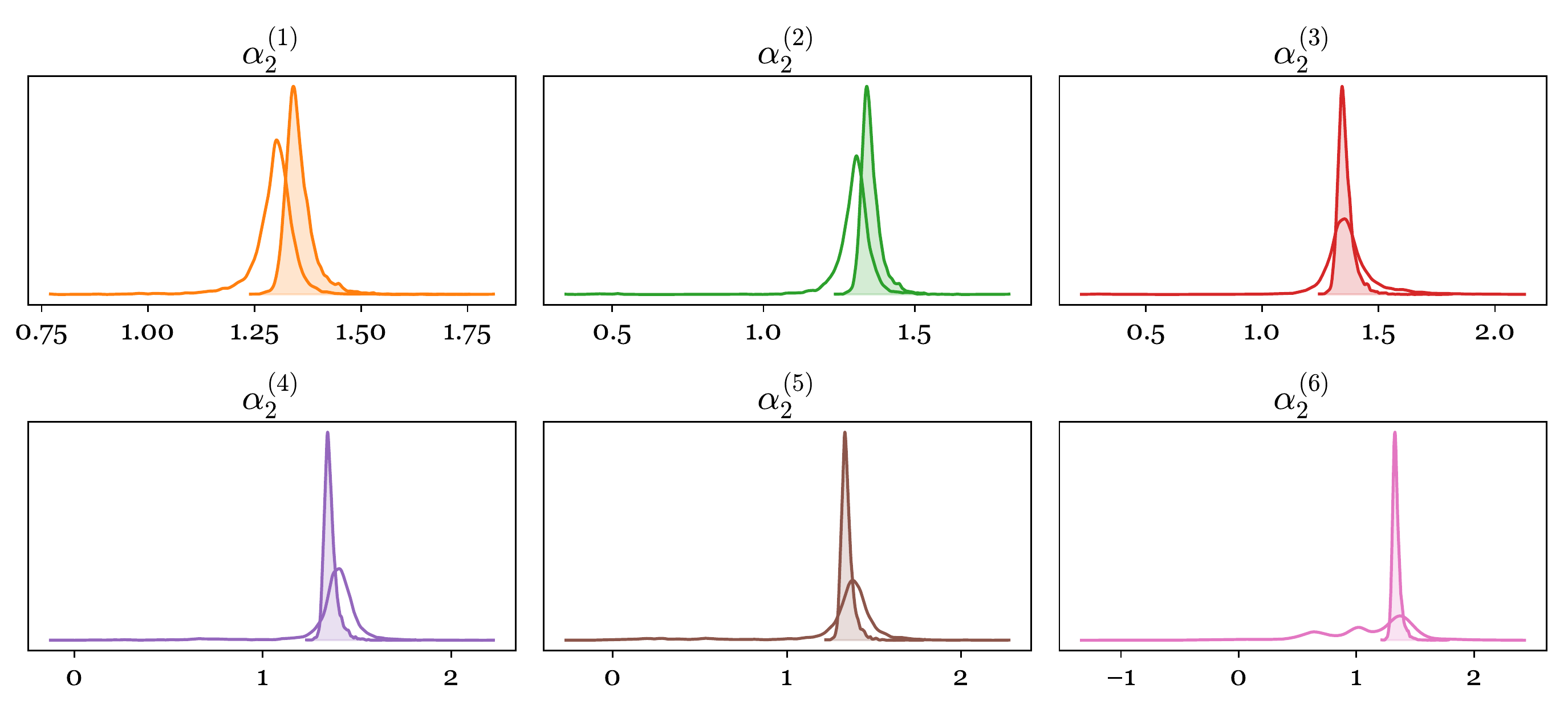}
    \caption{Variance reduction in the posterior distribution of the intercept $\alpha_1^{(k)}$ and slope $\alpha_2^{(k)}$ parameters for turbocharger components. Independent models (hollow) / population-level modelling (shaded).}
    \label{f:posts-alpha2TC}
\end{figure*}}

\section{\textit{Post publication note}}\label{a:post-pub}
Since publication, direct access to \textit{time at failure} observations ($T$) has revealed that subgroups in the empirical hazards are an artefact of the discreteness of the data. %
This does not detract from the demonstration, that knowledge transfer between fleet tasks is possible, however, in practice, the labels would not be connected to meaningful vehicle sub-groups (for this dataset).

\end{document}